\documentclass[lettersize,journal]{IEEEtran}
\usepackage{amsmath,amsfonts}
\usepackage{algorithmic}
\usepackage{algorithm}
\usepackage{array}
\usepackage[caption=false,font=normalsize,labelfont=sf,textfont=sf]{subfig}
\usepackage{textcomp}
\usepackage{stfloats}
\usepackage{url}
\usepackage{verbatim}
\usepackage{graphicx}
\usepackage{cite}
\usepackage{orcidlink}
\usepackage{lipsum}
\usepackage{makecell}
\usepackage{hhline}
\usepackage{mathtools}
\usepackage{caption}
\usepackage{subcaption}

\begin{document}

\title{On Model-Free Re-ranking for Visual Place Recognition with Deep Learned Local Features}

\author{Tom\'a\v{s} Pivo\v{n}ka$^{1, 2~ \orcidlink{0000-0002-1466-5156}}$  and Libor P\v{r}eu\v{c}il$^{1~\orcidlink{0000-0002-4286-3870}}$, ~\IEEEmembership{  Member,~IEEE}
\thanks{This work was co-funded by the European Union under the project Robotics and advanced industrial production (reg. no. CZ.02.01.01/00/22\_008/0004590).
}
\thanks{$^{1}$Czech Institute of Informatics, Robotics, and Cybernetics, Czech Technical University in Prague, Jugoslávských partyzánů 1580/3, 160 00 Praha 6, Czech Republic {\tt\small tomas.pivonka@cvut.cz}}
\thanks{$^{2}$Department of Cybernetics, Faculty of Electrical Engineering,
Czech Technical University in Prague, Karlovo náměstí 13, 121 35 Praha 2, Czech Republic}
}

\markboth{IEEE Transactions on Intelligent Vehicles, PREPRINT VERSION, ACCPETED MAY 2024 (DOI 10.1109/TIV.2024.3404564)}%
{Shell \MakeLowercase{\textit{et al.}}: A Sample Article Using IEEEtran.cls for IEEE Journals}

\IEEEpubid{\begin{tabular}[t]{@{}l@{}}© 2024 IEEE
Personal use of this material is permitted.  Permission from IEEE must be obtained for all other uses, in any current or future media,  including \\reprinting/republishing this material 
for advertising or promotional purposes, creating new collective works, for resale or redistribution to servers or lists, or \\reuse of any copyrighted component of this work in other works.\end{tabular}}

\maketitle

\begin{abstract}
Re-ranking is the second stage of a visual place recognition task, in which the system chooses the best-matching images from a pre-selected subset of candidates. Model-free approaches compute the image pair similarity based on a spatial comparison of corresponding local visual features, eliminating the need for computationally expensive estimation of a model describing transformation between images. The article focuses on model-free re-ranking based on standard local visual features and their applicability in long-term autonomy systems. 
It introduces three new model-free re-ranking methods that were designed primarily for deep-learned local visual features. These features evince high robustness to various appearance changes, which stands as a crucial property for use with long-term autonomy systems.
All the introduced methods were employed in a new visual place recognition system together with the D2-net feature detector (Dusmanu, 2019) and experimentally tested with diverse, challenging public datasets. 
The obtained results are on par with current state-of-the-art methods, affirming that model-free approaches are a viable and worthwhile path for long-term visual place recognition.
\end{abstract}


\section{Introduction}
\label{Introduction}

Visual place recognition (VPR) denotes the task of searching for the location from which a query image has been captured. The environment is represented by an image database of particular places or a model constructed from this database. In some cases, the global positions of individual database images are also known. Then, the output of the VPR can be the global position of a query image, its relative position to the best matching images, or just the index of the best matching image. This work is focused exclusively on the last of the herein listed types, searching for the best matching image without knowledge of global image positions. In this case, VPR is commonly approached as an image retrieval task.

In mobile robotics and autonomous vehicles, these VPR systems serve for basic localization and have two main applications. In simultaneous localization and mapping systems (SLAM), VPR systems support loop-closure detection, which triggers the global optimization of the map to reduce cumulative errors \cite{VPR_survey}. Secondly, the recognized image can be directly employed in autonomous vehicle navigation. For example, appearance-based teach-and-repeat systems directly control a~vehicle relying on a detected shift between a~query image and the closest image from the database built in the teaching phase~\cite{SSM-TaR}.

\IEEEpubidadjcol

Due to limited computational time, it is not usually possible to compare a query image with all images in a database in an~exhaustive manner. Therefore, VPR is usually split into two stages  - filtering and re-ranking \cite{VPR_survey}. In the filtering stage, an algorithm performs a fast search in the whole database and finds a selected number of best-matching candidate images. The found candidates are subsequently re-ranked by conducting a deeper comparison in the second stage of the VPR.

Spatial verification and non-geometric re-ranking belong to the major re-ranking approaches  \cite{VPR_survey}. Spatial verification methods use sparse local visual features matched between images to compute image similarity. These methods can further be categorized as model-based and model-free ones. \cite{Pairwise_match}.

Model-based systems estimate the spatial transformation between images, usually using robust methods to handle incorrect matches (e.g., RANSAC \cite{Patch-NetVLAD}), and evaluate similarity based on the model. Alternative model-free approaches directly compute image similarity from the matches, e.g., by histogram voting \cite{Zhang}.

Non-geometric methods do not rely on local correspondences and take advantage of other alternative principles.

\begin{figure}[t]
    \centering
    \includegraphics[width=0.49\linewidth]{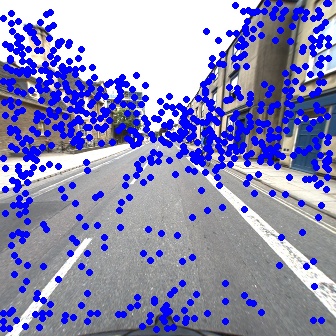}
    \hfill
    \includegraphics[width=0.49\linewidth]{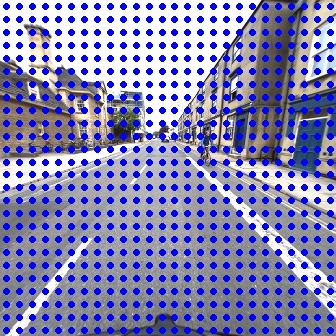}
    \caption{Example distributions of standard local visual features detected in an image structure (left) and features extracted at fixed grid positions (right)}
    \label{fig:features}
\end{figure}

So far, only few systems have used model-free approaches, especially in combination with deep-learned visual features \cite{VPR_survey}. The presented work was motivated mainly by the Semantic and Spatial Matching Visual Place Recognition system (SSM-VPR) \cite{SSM-VPR} that reached state-of-the-art results on many public datasets in 2020 and demonstrated the potential of model-free methods. However, this method is incompatible with standard local visual features and relies on its own type of local visual features that are extracted only at fixed, regular grid positions. Their main disadvantage is that these features do not directly relate to the image content structures. An~illustration of this concept compared with standard local visual features is provided in Fig. \ref{fig:features}.

The main goal of the presented work was to study and design new model-free approaches working with standard local visual features. In general, standard visual features can be better matched between images since they are related to truly existing structures in an image. They are also mostly detected in well-textured parts of an image. In addition, thanks to their universal applicability, other visual systems (e.g., visual odometry) can simultaneously re-use the same features, which saves computational resources in real applications. As this research focuses on methods for long-term autonomous systems, the new approaches were designed primarily for deep-learned local visual features that generally maintain high robustness to various appearance changes in the environment.

Here, we introduce three new model-free re-ranking methods designed for standard local features. For evaluation purposes, these re-ranking methods were tested with a robust D2-net feature detector \cite{D2-Net} that still belongs among state-of-the-art detectors for the VPR task \cite{local-features-for-VPR}. At first, the new re-ranking methods were combined with the filtering stage from the original SSM-VPR and tested on multiple challenging public datasets. These datasets contain various appearance changes in the environment imposed by moving objects, weather and seasonal variations, or diverse lighting conditions. Since the original SSM-VPR filtering stage proved to be limiting for the overall performance, the experiments for the best re-ranking method were repeated with the newer MixVPR system \cite{MixVPR} instead. In addition, a new approach combining filtering and re-ranking scores was also tested for this configuration.

The rest of the article is structured as follows: 
Sect.~\ref{RelatedWork} presents related work, including state-of-the-art VPR systems and existing model-free approaches to the re-ranking stage.
Subsequently, three hereby suggested model-free methods are presented in Sec. \ref{Methods}. 
The new VPR system using the introduced methods is described in Sect.~\ref{impl_details}.
The experimental evaluation and final discussion of the results follow in Sect.~\ref{Experiments}.
Finally, the paper is summarized in Sec.~\ref{Conclusions}.

\section{Related Work}
\label{RelatedWork}
\subsection{Visual Place Recognition}
\label{RW_VPR}
The crucial part of VPR is image representation. This representation has to be discriminative enough to distinguish between individual places but also sufficiently general to match images of the same place under different observation conditions. Most systems represent individual images by a~single fixed-length descriptor (e.g., \cite{VLAD}), so the images can be directly compared based on their descriptors' distances. These descriptors can be computed globally for the whole image or aggregated from multiple local descriptors (i.e., standard local visual features). Due to typically time-consuming computation, direct comparison of matched local features is usually applied in the re-ranking stage only \cite{VPR_survey}.

Major current state-of-the-art VPR methods are based on convolutional neural networks (CNNs). There are two main ways to create a global descriptor from the convolutional layer output \cite{VPR_survey}. Systems with aggregated representations treat individual columns of an output tensor similarly to local features and use them to create the final descriptor. The second approach relies on pooled representations whose descriptors are directly composed of pooling operations (e.g., max-pooling) applied to individual feature maps of the output tensor.

Aggregated representation is used by the NetVLAD system \cite{NetVLAD} composed from two different neural networks. The first network is part of a standard classification neural network. An~output of its mid-layer is used to extract local visual features as columns of its output tensor. The second network has a special architecture inspired by VLAD \cite{VLAD} and serves to aggregate all features into the final descriptor. It has achieved state-of-the-art performance on many datasets, and many current systems proceed from this solution.

In \cite{HDME}, NetVLAD is extended by a re-ranking stage using SuperPoint visual features \cite{SuperPoint} and SuperGlue matcher \cite{SuperGlue} to find correspondences between images. The similarity of two images is measured as the count of in-lying matches for a homography between images estimated by a model-based method. In addition, the method filters out features based on evaluation maps computed from VLAD clusters.

Another extension of NetVLAD is Patch-NetVLAD \cite{Patch-NetVLAD}. Similarly to \cite{HDME}, it uses the original NetVLAD to select the best-matching candidates in the filtering stage. In the re-ranking stage, the method extracts NetVLAD descriptors for small image patches at various scales. These patches are matched between images equally as local features. Similarity is measured by two different approaches. The first is a model-based approach that computes a~homography. The second one relies on a model-free method, estimating similarity directly from the horizontal and vertical displacement of features that are matched between images. The experiments presented in \cite{Patch-NetVLAD}  show only a negligible decrease in the performance of the model-free approach compared to the model-based one.

A general approach to pooling representation named generalized mean pooling (GeM) is presented in \cite{GeM}. It combines maximum and average pooling with a learnable parameter for each feature map. In \cite{CosPlace}, the GeM layer is followed by a fully connected layer to get the final descriptor. It achieved state-of-the-art results on many datasets in combination with introduced CNN training techniques designed for large datasets.

Another system using pooling representation is MixVPR \cite{MixVPR}. At first, the system uses Feature-Mixer, which is a cascade of fully connected layers applied to individual flattened feature maps of a CNN output tensor. The shape of the mixer output is equal to flattened maps stacked into a 2D matrix. Therefore, its dimension is further reduced by two 1D fully connected layers (weighted pooling) applied in horizontal and vertical directions. The reduced matrix is further normalized and flattened to obtain the final global descriptor. Although MixVPR does not use the re-ranking stage, it achieves a similar level of precision as the current state-of-the-art systems.

The last mentioned system is SelaVPR \cite{SelaVPR}, reaching the current state-of-the-art results on many datasets. This system is based on vision transformers that currently outperform CNN approaches in many computer vision tasks. SelaVPR reshapes the patch tokens computed by the vision transformers from the input image to a 3D tensor similar to CNN output. A global descriptor for the filtering stage is created directly by applying GeM to this tensor. By contrast, local features are extracted as columns of this tensor up-sampled by the Local Adaptation Module. The used model-free re-ranking approach computes image similarity directly as the count of the mutual nearest neighbor matches.

\subsection{SSM-VPR}
\label{RW_SSM}
Semantic and Spatial Matching Visual Place Recognition system \cite{SSM-VPR} (mentioned in Sect. \ref{Introduction}) follows the standard two-stage VPR pipeline. The main contributions include an approach to extracting local visual features from CNNs and a~model-free method for spatial comparison in the re-ranking stage. In addition, this system reaches state-of-the-art results with many public datasets \cite{SSM-VPR}.

Visual features are extracted in the form of sub-cubes from an output tensor of the selected CNN inner layer. The method does not search for any key points and directly creates the features at fixed square grid positions covering the whole image. Each sub-cube is transformed into a one-dimensional vector, and its size is reduced using principal component analysis (PCA) to obtain the final descriptor. The filtering stage uses visual features extracted from higher CNN layers that contain substantial semantic information. In the re-ranking stage, the features are extracted from lower layers instead, providing more accurate spatial information.

The re-ranking method presented in \cite{SSM-VPR} is based on the concept of anchor points. It is described together with our proposed version modified for standard local features in Sect.~\ref{Anchor_points}.

\subsection{Model-Free Re-ranking Approaches}
\label{RW_model-free}
Most of the model-free approaches apply generalized Hough transform \cite{Lowe2004} to measure the similarity between images. The main principle is a voting mechanism where each feature correspondence votes in a multi-dimensional histogram of possible transformations. The method presented in \cite{Zhang} uses a~2D histogram to represent translations along the $x$ and $y$ axes. In \cite{Avrithis}, authors use a 4D space including rotations and scales provided by local features in addition to their translations. The space is divided hierarchically, and the similarity is computed from grouped matches in bins through all levels. The 4D space is also used in a special version of VPR, searching only for a~selected region of an image \cite{Shen}. This approach computes voting maps for multiple rotated and re-scaled versions of the region. Here, each matched feature votes around an estimated center of the region using a 2D Gaussian weight function. The final best match is determined as a maximum value from all voting maps.

While all mentioned methods estimate transformation from single matches, a method using pairwise matching is presented in \cite{Pairwise_match}.

Another possible approach is to compute image similarity from all matches directly. This method is proposed in the aforementioned Patch-NetVLAD article \cite{Patch-NetVLAD}, where similarity is computed as a sum of residual displacements from the mean along the $x$ and $y$ axes subtracted from the maximum possible spatial offset. A similar method is also proposed in \cite{LoST}, where the best matching candidate has the lowest score computed as a sum of differences along the x-axis multiplied by a distance of descriptors for all matches.

\section{Model-Free Re-ranking Methods}
\label{Methods}
This section presents three new model-free re-ranking methods. These methods re-rank candidate images (selected in the VPR filtering stage) based on computed similarity scores to the query image and return the final best-matching image. The similarity score of an image pair is a scalar value computed from mutual correspondences of local visual features.

The inputs of the presented methods are positions of detected local visual features and their correspondences that match the most similar visual features between images. Optional inputs are quality scores of detected features or distances of matched descriptors, both used for weighting the individual matches.

 \subsection{Histogram of Shifts}
\label{Histogram}
The first method computes a histogram of shifts between matched features (i.e., differences of their positions). Similarly to \cite{Zhang}, a 2D histogram with square bins of defined size is created for all possible shifts along the $x$ and $y$ axes. Subsequently, each match votes in the histogram. The main difference from the previous works cited in Sect. \ref{RW_model-free} is that each shift votes via a Gaussian distribution to all bins, not only for the directly corresponding one. The voting mechanism is expressed in the following equation that describes the computation of a particular bin score.

\begin{equation}
    s_{i, j} = \sum_{k=1}^{n} w_k \cdot
    exp \left( -\frac{(x_k-x_{i,j})^{2}+(y_k-y_{i,j})^{2}}{2 \cdot \sigma^{2}} \right)
\end{equation}

The bin score $s_{i,j}$ is computed using all $n$ matches with shifts $x_k$ and $y_k$. These shifts are differences in the $x$ and $y$ coordinates of matched features. $x_{i,j}$ and $y_{i,j}$ are values of shifts that correspond to the bin center. $w_k$ is a~scalar representing a match weight. Specific weight metrics are presented in Sect. \ref{MatchWeights}. The last static parameter $\sigma$ determines the variance of the Gaussian distribution.

The final similarity score is selected as the maximum value from all bins. Besides, the method directly returns a dominant shift, which is usable for simple robot navigation mentioned in Sect. \ref{Introduction}.

\begin{figure}[t]
    \centering
    \includegraphics[width=\columnwidth]{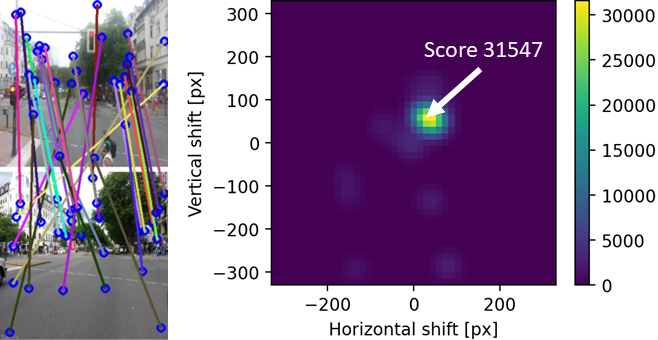} 
    \caption{
    Matched local visual features between images and the corresponding histogram of shifts (the number of matches was limited to 30)
    }
    \label{fig:histogram}
\end{figure}

Fig. \ref{fig:histogram} shows an example of the histogram with the highlighted final score and the corresponding image pair with depicted feature matches.

The main advantage of histogram voting is the ability to suppress incorrect matches. However, if the corresponding images are not captured at exactly the same place, even correctly matched features do not always exhibit the same shift due to the projection geometry. The Gaussian voting was introduced mainly to minimize the influence of this effect without the need to enlarge bin size. Appropriate weighting of the matches increases the impact of high-quality matches.

\subsection{Anchor Points}
\label{Anchor_points}
The second method was designed to merge the successful approach of SSM-VPR \cite{SSM-VPR} based on the anchor points principle with standard local visual features related to the image structure. A crucial property of the original method is that visual features are extracted at fixed square grid positions (Fig.  \ref{fig:features}), so feature matches between two images can be stored in a~2D matrix. This matrix corresponds to the spatial arrangement of visual features in the square grid. Each matrix element represents the specific visual feature from the first image and stores the position of the most similar feature with the closest descriptor in the second image. This position is represented by a unique index of the particular place in the grid that can be decomposed into two coordinates in the 2D grid of features.

This matrix representation is not applicable to standard local visual features since they are irregularly distributed across the images, and their count fluctuates. However, their matches can be transformed into a similar representation with fixed positions of features and then processed almost equally. This approach is introduced below.

The first image is divided into square bins emulating the grid of original SSM-VPR features, and the local features are organized into groups corresponding to particular bins. In each bin, the assigned features vote for the corresponding bin in the equally divided second image using positions of their matches. The voting strength of each match is determined by its weight. The bin with the highest score is selected as the final match. In this manner, the local features are transformed into a matrix representation almost equal to the original SSM-VPR. The only difference is that not all the bins necessarily have a match.

\begin{figure}[t]
    \centering
    \includegraphics[width=\columnwidth]{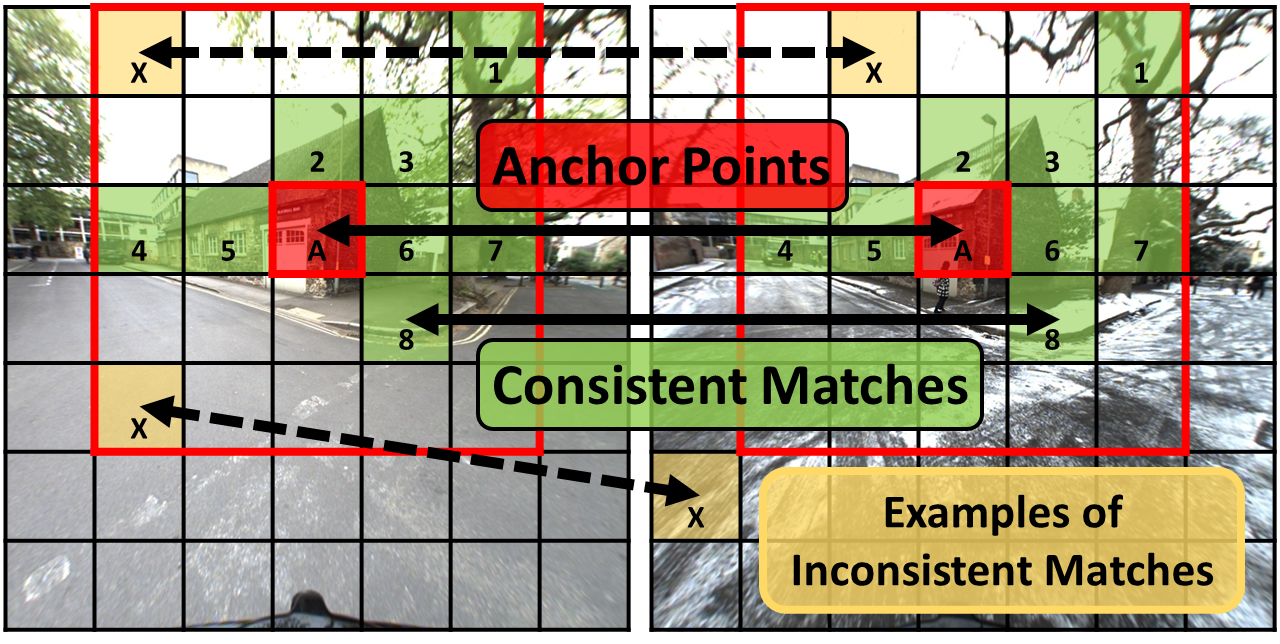}
    \caption{Anchor points principle. Individual cells represent local visual features extracted at fixed grid positions. Green cells belong to neighboring matches consistent with the selected anchor points pair (red).}
    \label{fig:anchor_points}
    \vspace{-5mm}
\end{figure}

The original concept of anchor points and similarity score computation is visualized in Fig. \ref{fig:anchor_points}. A pair of anchor points is a selected match between the two images. The model-free method checks if the relative positions of neighboring features and their matches to the anchor points from the pair are equal in both images, i.e., mutually consistent. The neighboring features are selected from a square window around the anchor point in the first image. The score of each anchor point is the count of consistent features, and the final candidate similarity score is the sum of scores for all features taken as anchor points.

Here, the following adjustments to the original anchor points method are introduced. Bins without a match are skipped during anchor points selection and considered inconsistent matches otherwise. The newly introduced shift tolerance relaxes the conditions for a consistent match. All matches within a threshold distance from a consistent position are included in the score computation, with the weight equal to the difference between the threshold and the distance. The score computation is also extended to the weighting by the bin score used in the conversion to matrix representation.

Score $s_{i, j}$ for one pair of anchor points, represented by an~element at position $i$, $j$ in the matrix of matches, is computed as:

\begin{equation}
    s_{i, j} = w_{i,j} \cdot \sum_{\{k,l\} \in N } max \left( 
     t - \left| \left| \begin{multlined} \left( \left[ \begin{array}{c} i \\ j \\ \end{array} \right] -  \left[ \begin{array}{c} k \\ l \\ \end{array} \right] \right) \\
    - (\vec{m}_{i,j} - \vec{m}_{k,l}) \end{multlined} \right| \right| , 0  \right)
\end{equation}

$w_{i, j}$ is a weight of a bin match. $N$ is a set of matrix elements with coordinates ($k$, $l$) in a square window around the anchor point element at position $i$, $j$. Parameter $t$ is a~tolerance threshold distance. $\vec{m}_{i,j}$ and $\vec{m}_{k,l}$ are the coordinates of corresponding bins in the second image to bins at positions ($i$, $j$) and ($k$, $l$). All bin coordinates are integers denoting position in a grid. $||.||$ denotes the Euclidean norm of a vector.

\subsection{Aggregated Score}
\label{Aggregation}
The last herein-developed method aggregates the similarity score from all matches directly. It is derived from the model-free approach presented in the Patch-NetVLAD system \cite{Patch-NetVLAD}. In comparison with the original method, we introduce several improvements. The maximum possible shifts are set to equal image sizes instead of the maximum measured shifts\footnote{In \cite{Patch-NetVLAD}, even though the value is referred to similarly as a maximum possible shift, it is expressed as a maximum value from all measured shifts in the equation. The usage of the maximal measured shifts does not work properly, as it rewards the images with the largest outlier.}.
The second adjustment is an introduction of the match weight to the computation. The last change is the removal of the sum normalization by the number of features. The equation for a~computation of candidate similarity score $s$ from shifts $x_k$, $y_k$  between matched features along the $x$ and $y$ axes is:

\begin{equation}
    s = \sum_{k=1}^{n} w_k \cdot \left(
        \left(x_{im} - |x_k - \overline{x}|  \right)^{2} + 
        \left(y_{im} - |y_k - \overline{y}|  \right)^{2} 
        \right)
\end{equation}

$x_{im}$ and $y_{im}$ are the image width and height, $\overline{x}$ and $\overline{y}$ are the mean values from all the shifts, and $w_k$ is the match weight.

Here, the primary advantage of the approach is a low computational cost and substantially higher robustness to outliers compared to a direct sum of weights of all matches.

\begin{figure}[t]
    \centering
    \includegraphics[width=\columnwidth]{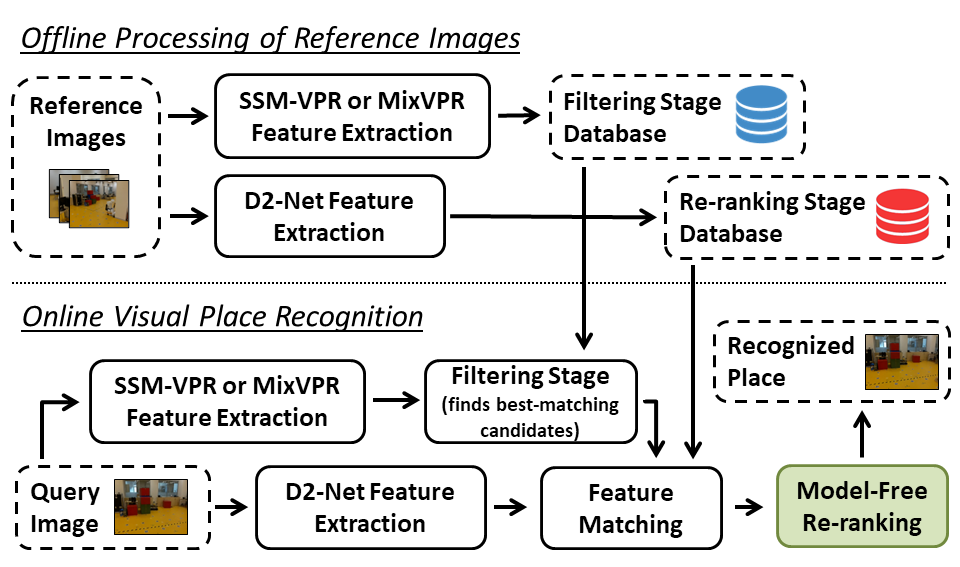}
    \caption{The new VPR system with the SSM-VPR or MixVPR filtering stage and re-ranking stage using the introduced model-free methods and D2-Net features.}
    \label{fig:flowchart}
\end{figure}

\section{Implementation Details}
\label{impl_details}
\subsection{Visual Place Recognition System}
\label{VPR_system}

The introduced model-free methods were implemented into a new VPR system. The system combines the SSM-VPR filtering stage \cite{SSM-VPR} or MixVPR \cite{MixVPR} with the new re-ranking methods. Local visual features in the second stage are extracted by the D2-net detector \cite{D2-Net}. This represents a state-of-the-art deep-learning-based solution \cite{local-features-for-VPR} that excels in robustness to significant appearance changes. Visual features are extracted from CNN layers similarly to SSM-VPR, but they are related to the structure of the image and not detected at fixed positions. Therefore, features can be matched more precisely, and they are especially detected in the well-textured image parts. In the presented system, the features are extracted on a single image scale by a default network trained by the D2-Net authors on the MegaDepth dataset \cite{MegaDepth}.

The flowchart of the whole system is shown in Fig.~\ref{fig:flowchart}. Visual feature detection is performed for both stages separately, as well as the database creation. The re-ranking stage database stores D2-net features for all reference images. During the recognition, D2-net features from the query image are matched with the features of the best-matching candidates returned by the filtering stage. The system uses a~standard nearest neighbor matcher with cross-check based on the Euclidean distance of descriptors. Cross-check is an additional step that keeps only matches with mutually nearest features. Afterward, the selected model-free method computes a similarity score for each candidate image from the matches. Finally, the image with the highest score is returned as the best image representing the recognized place.

\subsection{Weights of Matches}
\label{MatchWeights}
All model-free methods introduced in Sect. \ref{Methods} use weights of matches to compute the similarity score. In the presented VPR system, two types of weights are used.

The first approach, termed FS-weight, takes advantage of the D2-net detector \cite{D2-Net} that returns the score for each feature expressing its quality. The match weight is then calculated as a sum of both feature scores in a match ($w = s_1 + s_2$).

The second type (DMF-weight) uses distances between matched feature descriptors. 
Since the values of distances between L2 normalized descriptors lie in the interval $[0, 2]$ ($d \in [0, 2]$), the weight can be determined as $w = 2 - d$.

\subsection{Parameters of Model-Free Methods}
\label{parameters}

The performance of all presented methods depends on the proper setting of their inner parameters. Below are summarized all optional parameters and their values used for the experiments presented in Sect. \ref{Experiments}. These values were selected based on preliminary tests performed on three sequences of the RobotCar dataset \cite{RobotCar1, RobotCar2}. The presented values correspond to the used image sizes 336 x 336 px. The optimal values for other image sizes can be different.

The inner parameters for the histogram of shifts method and their optimal values are the size of histogram bins = 15px, the variance of the Gaussian = 22.5, and the type of match weight = FS-weight. This method, using the SSM-VPR filtering stage, is going to be further referred to as the SSM-hist.

The anchor points method (SSM-anchor) has the following inner parameters and settings: number of bins = 15, size of the window = 10, tolerance threshold = 3, and type of match weight = FS-weight.

The only parameter for the aggregated score method (SSM-sum) is the type of match weight, and the selected approach was DMF-weight.

\subsection{Reference Model-Based Method}
\label{Model-based-system}
The RANSAC homography estimation method presented in \cite{local-features-for-VPR} was another re-ranking approach implemented into the presented system for experimental comparison of model-free and model-based approaches. The similarity score is determined by the count of consistent matches for the best-found model. In this configuration, termed SSM-RAN, the system uses the SSM-VPR filtering stage and D2-net feature detector similarly to configurations introduced in Sect. \ref{parameters}.

The used re-ranking image size is 336 x 336 px, being consistent with model-free methods. The optional parameters for the RANSAC estimation are the maximum number of iterations (set to 2000) and the reprojection threshold. The system was tested with two different threshold values - 24 px (the original value from \cite{local-features-for-VPR}) and 8~px (value proportionally adjusted to the used resolution).

\subsection{MixVPR and Combined Re-ranking Score}
\label{MixVPR-version}
The histogram of shifts method, which achieved the best performance among the proposed methods (see Sect. \ref{Experiments}), was also combined with the MixVPR system \cite{MixVPR} instead of the SSM-VPR filtering stage. This system version is termed Mix-hist. It uses a default image resolution for MixVPR 320~x~320~px. Since it is similar to the resolution of SSM-hist, the values of inner parameters were set equally to parameters introduced in Sect. \ref{parameters}. The system uses the original version of MixVPR using ResNet50 backbone and feature dimension 4096, which was pre-trained on the GSV-Cities dataset \cite{MixVPR}.

Since MixVPR achieves high performance itself, the score from the filtering stage could also be valuable for re-ranking. A modified version of Mix-hist combining both scores termed Mix-hist-comb was created to test this hypothesis. The filtering score $s_f$ and re-ranking score $s_r$  were combined directly as:

\begin{equation}
    s = c \cdot s_f + s_r
\end{equation}

$c$ is a fixed constant set to ensure a comparable influence of both $s_f$ and $s_r$ values. For the presented experiments, the value typically ranges around $10^6$.

\section{Experiments}
\label{Experiments}

\subsection{Datasets}
\label{Datasets}
The new VPR system with integrated, presented model-free methods was evaluated with six public datasets listed below. Their major properties are summarized in Tab. \ref{Tab3}.

\begin{table}[b]
  \caption{Used datasets and their properties (number of reference and query images and types of appearance changes)}
    \centering
    \renewcommand{\arraystretch}{1.3}
    \begin{tabular}{|l|c|c|c|c|}
    \hline
     & \makecell{ Ref. \\ Images } & \makecell{ Query \\ Images } & \makecell{ Day/Night \\ Changes } & \makecell{ Seasonal \\ Changes } \\ \hline
    Berlin A100 & 85 & 81 & No & No \\ \hline
    Berlin Halenseestrasse & 67 & 157 & No & No \\ \hline
    Berlin Kudamm & 201 & 222 & No & No \\ \hline
    Gardens Points & 200 & 200 & Yes & No \\ \hline
    Synthesized Nordland & 1415 & 1415 & No & Yes \\ \hline
    Oxford RobotCar & 220 & 1411 & Yes & Yes \\ \hline
    \end{tabular}
  \label{Tab3}
\end{table}

\emph{Berlin A100} \cite{Mapillary, Berlin-datasets} is a small dataset captured at an~urban motorway from two different cars. Both sequences were taken in the daytime. The dataset is challenging mainly due to significant viewpoint changes.

\emph{Berlin Kudamm} \cite{Mapillary, Berlin-datasets} dataset was captured at a busy urban street in the daytime from a bus as a reference and a bike for query images. Besides a significant viewpoint difference, it also contains many dynamic objects, such as cars or pedestrians.

\emph{Berlin Halenseestrasse} \cite{Mapillary, Berlin-datasets} is an urban dataset composed of two image sequences captured by car and bike users. Similarly to Berlin A100 and Kudamm, both sequences were taken in the daytime. The dataset is challenging due to large viewpoint changes caused mainly by the distance between the roadway and the bike lane, even separated by vegetation in some places.

\emph{Gardens Points} \cite{GardensPoint} contains two traversals captured in the daytime and by night in a university campus that mix indoor and outdoor environments. The night sequence contains only noisy black-and-white images, which makes stable feature matching difficult.

\emph{Synthesized Nordland} \cite{NordLand, NordLandPaper} was captured from a train in Norway during summer and winter. The images contain mainly seasonal changes in the presence of snow and the look of vegetation.

All the aforementioned datasets\footnote{available at http://imr.ciirc.cvut.cz/Datasets/Ssm-vpr} were adopted from \cite{SSM-VPR}.

\emph{Oxford RobotCar} \cite{RobotCar1, RobotCar2} is a public dataset for long-term visual localization. It contains many image sequences captured from a car driven along the same trajectory in an~urban environment at various weather conditions, daytimes, and seasons. The dataset was originally created for full 6DoF localization. It provides the ground truth positions for approximately half of the images. For the presented experiments, only the images with known ground truth were used and matched between sequences applying the closest distance objective with the removal of matches distant more than 5 m. The overcast-reference sequence was used for reference images and matched to the other seven sequences (dawn, dusk, night, night-rain, overcast-winter, rain, snow).

\begin{figure*}[ht]
    \centering
    \includegraphics[width=\textwidth]{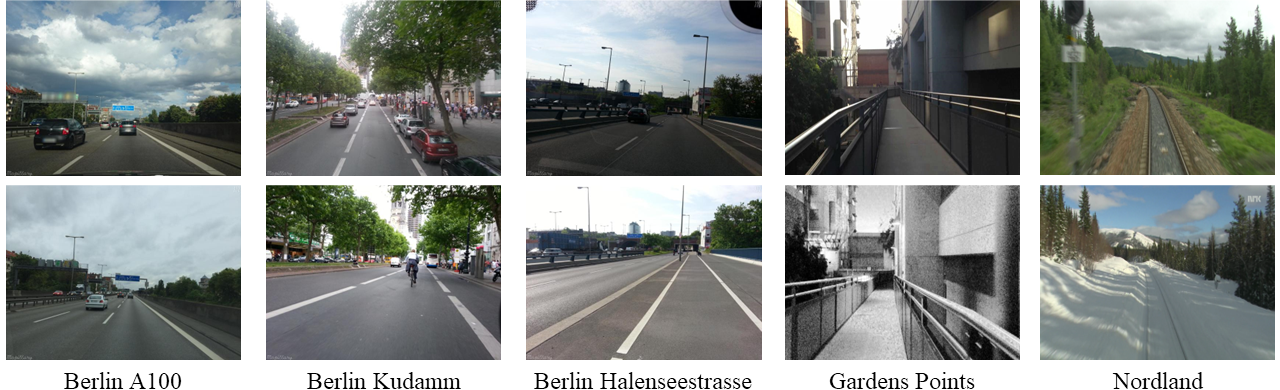}
    \caption{Example images from databases introduced in \cite{SSM-VPR} and Sect. \ref{Datasets}.
    The upper row contains images from reference sequences and the bottom row depicts corresponding images from query sequences.}
    \label{fig:datasets1}
\end{figure*}

\begin{figure*}[ht]
    \centering
    \includegraphics[width=\textwidth]{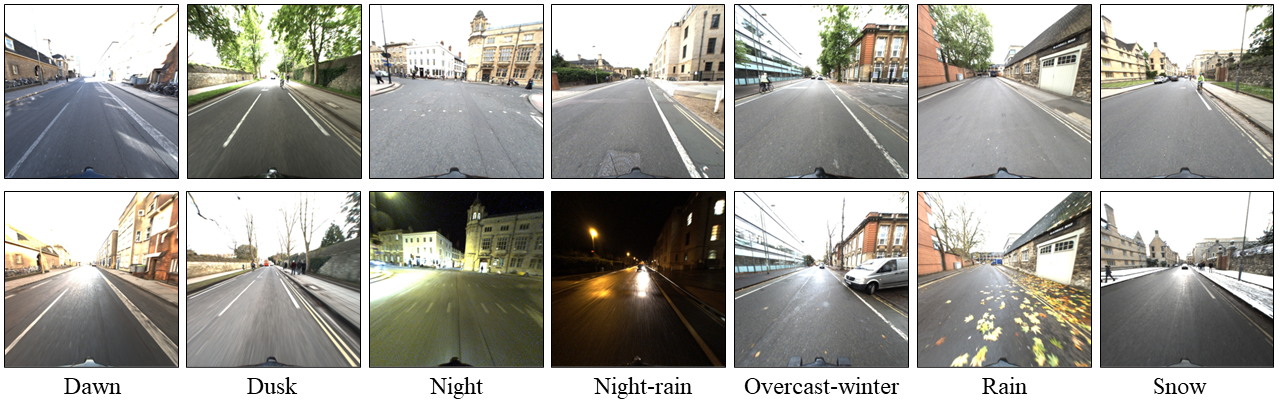}
    \caption{Examples of images from all RobotCar dataset's query sequences used for testing and their corresponding database images from overcast-summer sequence in the upper row. \cite{RobotCar1, RobotCar2}.}
    \label{fig:datasets2}
\end{figure*}

Example images of all datasets are shown in Fig. \ref{fig:datasets1} and \ref{fig:datasets2}.

\subsection{Experimental Setup and Results}
The presented model-free methods combined with the SSM-VPR filtering stage were tested with all datasets introduced in Sect. \ref{Datasets} and directly compared to the selected reference systems. The histogram of shifts method was also complemented with MixVPR in the filtering stage, as introduced in Sect.~\ref{MixVPR-version}. The reference systems were the original SSM-VPR \cite{SSM-VPR}, the new system SSM-RAN with model-based re-ranking (Sect.~\ref{Model-based-system}), two basic versions of Patch-NetVLAD \cite{Patch-NetVLAD}, MixVPR \cite{MixVPR} and SelaVPR \cite{SelaVPR}.

The first version of Patch-NetVLAD, termed Patch-NV perf., is optimized for the best performance. It uses multilevel feature detection and model-based re-ranking based on RANSAC. The second version optimized for speed (Patch-NV speed) employs only single-layer feature detection and the model-free re-ranking mentioned in Sect. \ref{RW_model-free}. The used image sizes for the Patch-NetVLAD were 480 x 480 px for the RobotCar dataset and 640 x 480~px for the others. The number of candidates returned by the filtering stage was set to 100, which is the default value provided by the authors of Patch-NetVLAD.

\begin{table}[t]
    \centering
    \caption{VPR systems setting for the experimental evaluation}
    \renewcommand{\arraystretch}{1.3}
    \begin{tabular}{|l|c|c|c|}
    \hline
     & \makecell{ Filtering \\ Resolution } & \makecell{ Re-ranking \\ Resolution } & \makecell{ Number of \\ Candidates } \\ \hline
    Patch-NV & \makecell{ 640 x 480 \\ 480 x 480 } & \makecell{ 640 x 480 \\ 480 x 480 } & 100 \\ \hline
    MixVPR & 320 x 320 & - & - \\ \hline
    SelaVPR & 224 x 224 & 224 x 224 & 50\\ \hline
    SSM-VPR & 224 x 224 & 336 x 336 & 50\\ \hline
    SSM-hist, anchor, sum, RAN  & 224 x 224 & 336 x 336 & 50 \\ \hline
    Mix-hist & 320 x 320 & 320 x 320 & \makecell{10 \\ 50} \\ \hline
    \end{tabular}
  \label{Tab_params}
\end{table}

Used image sizes and the number of candidates for other methods are presented in Tab. \ref{Tab_params}. Mix-hist and Mix-hist-comb systems were tested for two different numbers of candidates, 10 and 50 candidates. The higher number of candidates increases the influence of the re-ranking stage on the overall system performance. Two diverse versions of the system were also used for SelaVPR and SSM-RAN. SelaVPR was tested with two different models pre-trained on Pittsburgh30k (SelaVPR pits.) or Mapillary Street-level Sequences (SelaVPR msls) datasets. SSM-RAN versions differed by the re-projection threshold value (8 or 24 px).

The used metric is Recall@1, evaluating the percentage of correctly recognized places for the best matches only. All experiments were performed with three various tolerance ranges. The tolerance range is a distance from the ground truth image in a reference image sequence, in which the recognized place is still accepted as correct. The chosen values were 1, 2, and 5 images for all the used datasets.

\begin{figure}[t]
    \centering
    \includegraphics[width=\columnwidth]{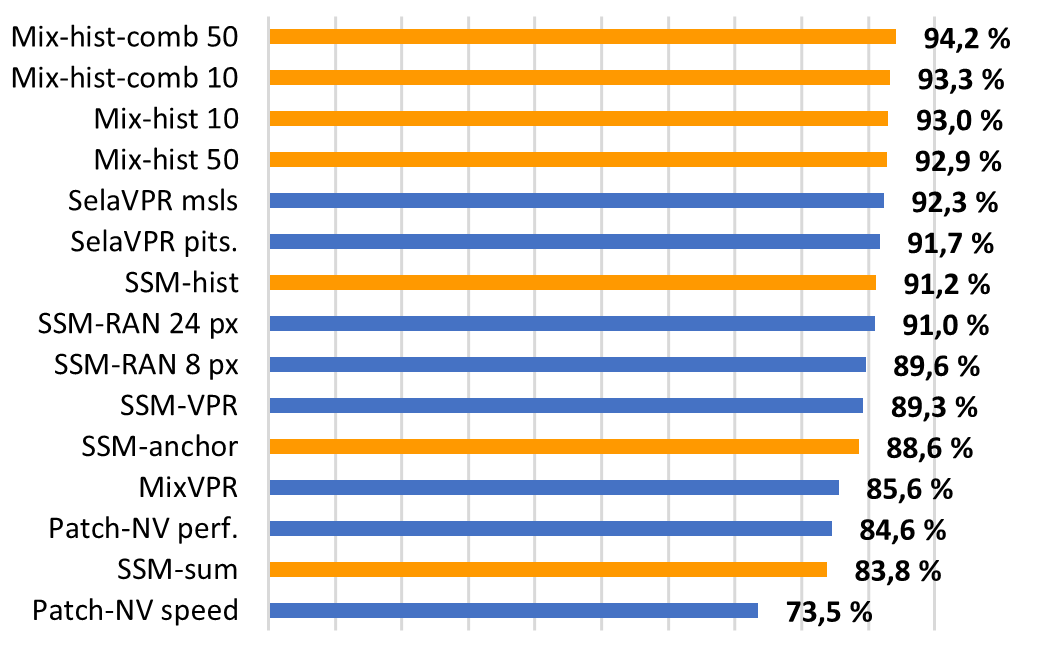}
    \caption{Average Recall@1 [\%] in all datasets with tolerance range 2. Orange bars are results of the presented systems and blue bars are results of reference systems.}
    \label{fig:sum}
\end{figure}

\begin{figure*}[hp]
    \centering
    \subfloat[Berlin A100]{
    \includegraphics[width=0.49\linewidth]{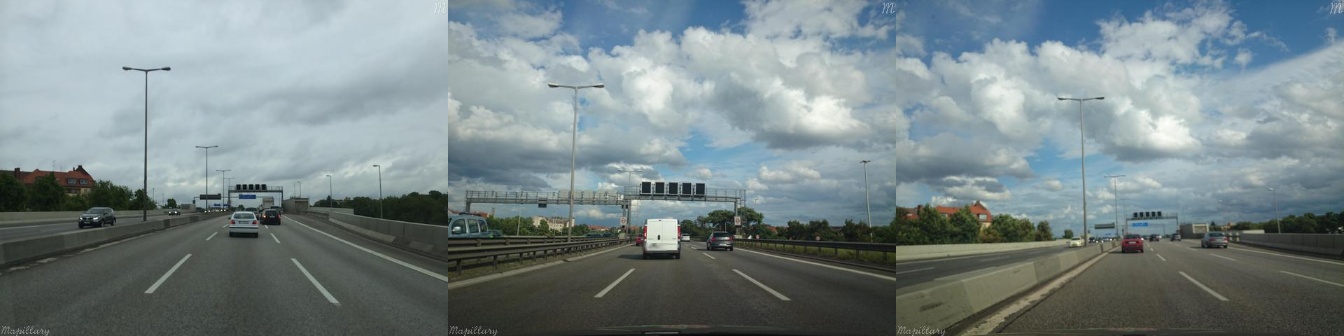}
    \hfill
    \includegraphics[width=0.49\linewidth]{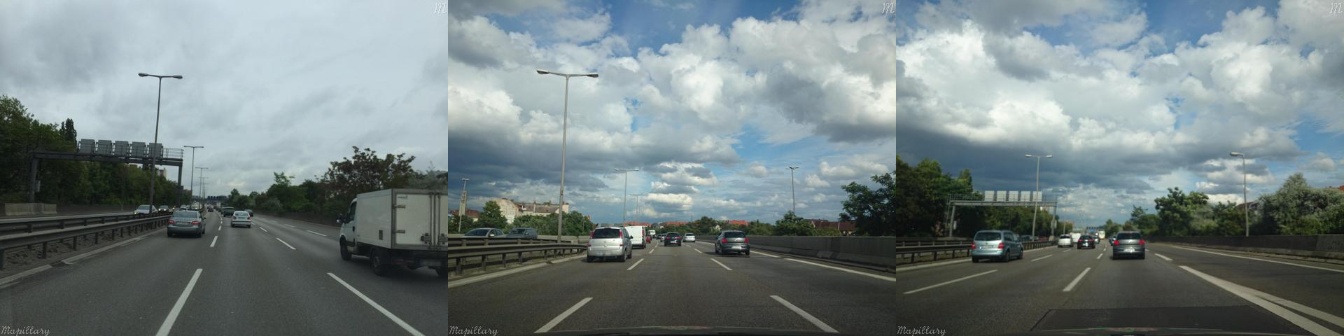}
    }
    
    \subfloat[Berlin Kudamm]{
    \includegraphics[width=0.49\linewidth]{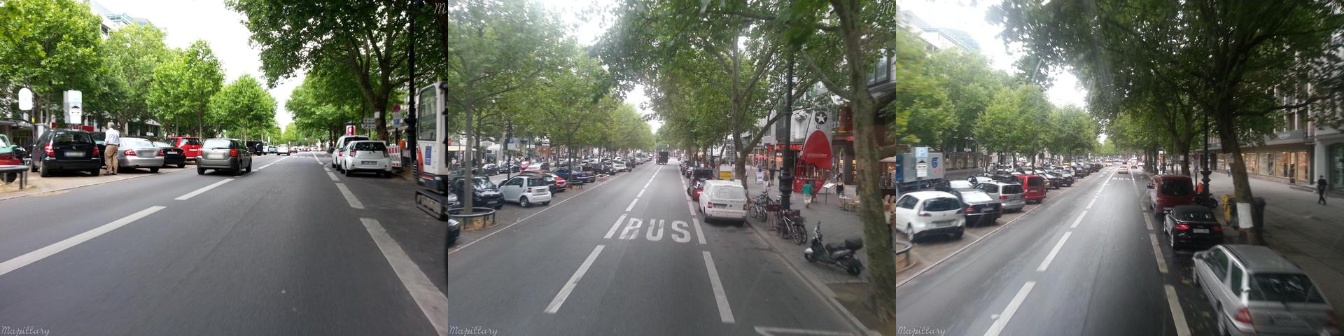}
    \hfill
    \includegraphics[width=0.49\linewidth]{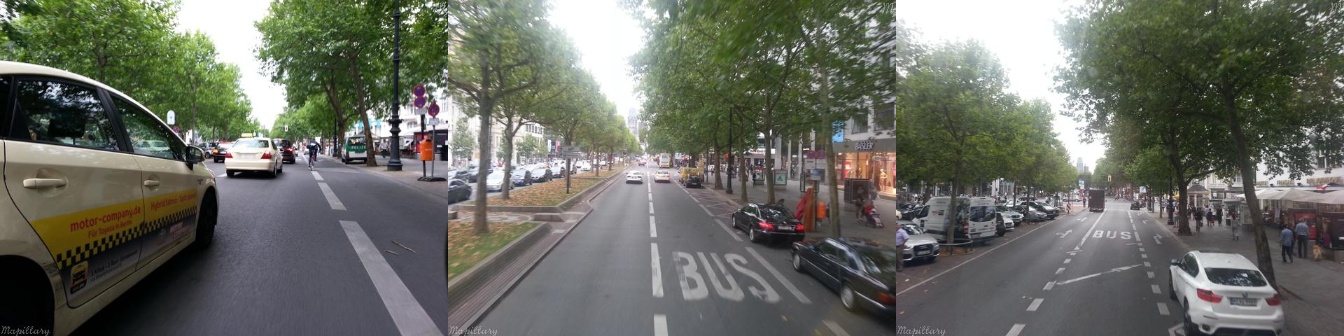}
    }
    
    \subfloat[Berlin Halen.]{
    \includegraphics[width=0.49\linewidth]{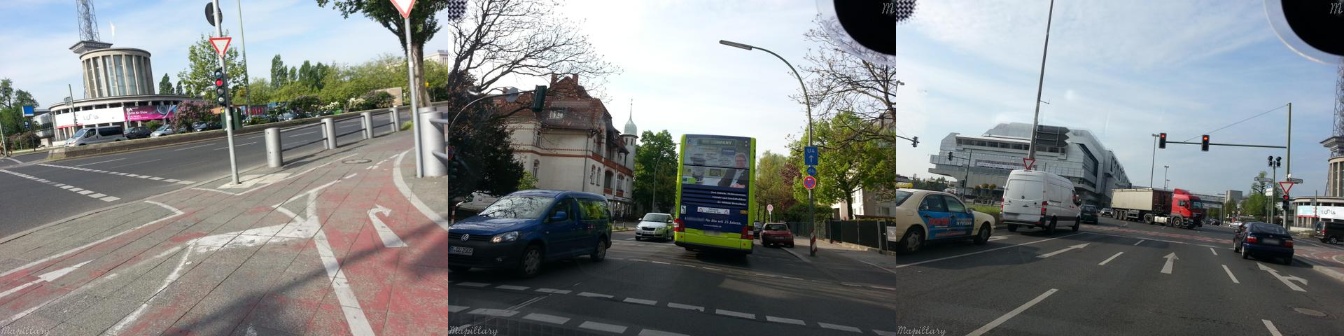}
    \hfill
    \includegraphics[width=0.49\linewidth]{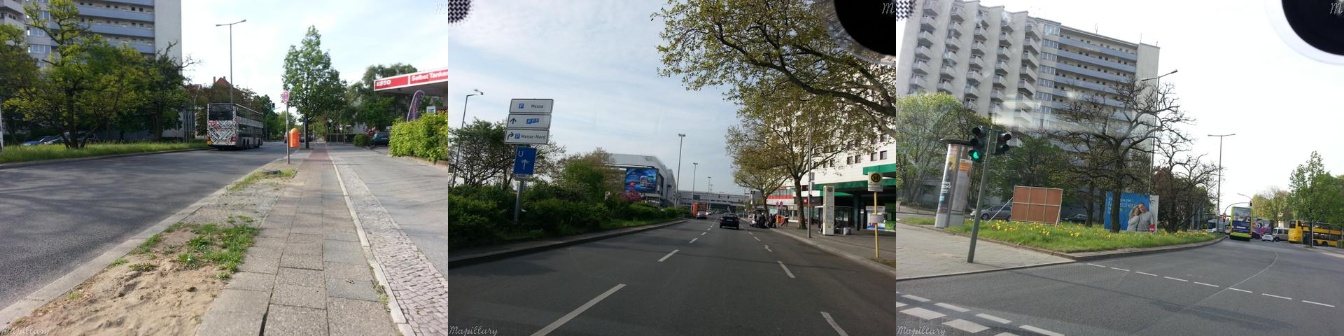}
    }

    \subfloat[Gardens Points]{
    \includegraphics[width=0.49\linewidth]{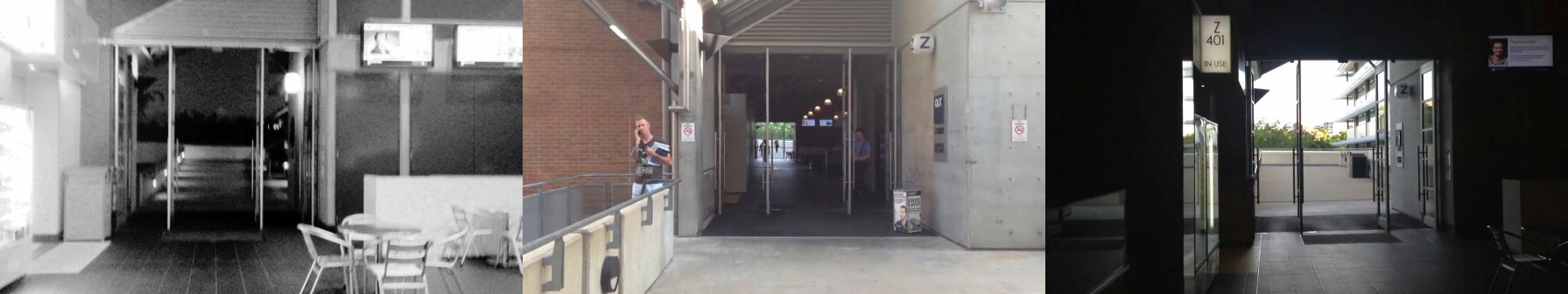}
    \hfill
    \includegraphics[width=0.49\linewidth]{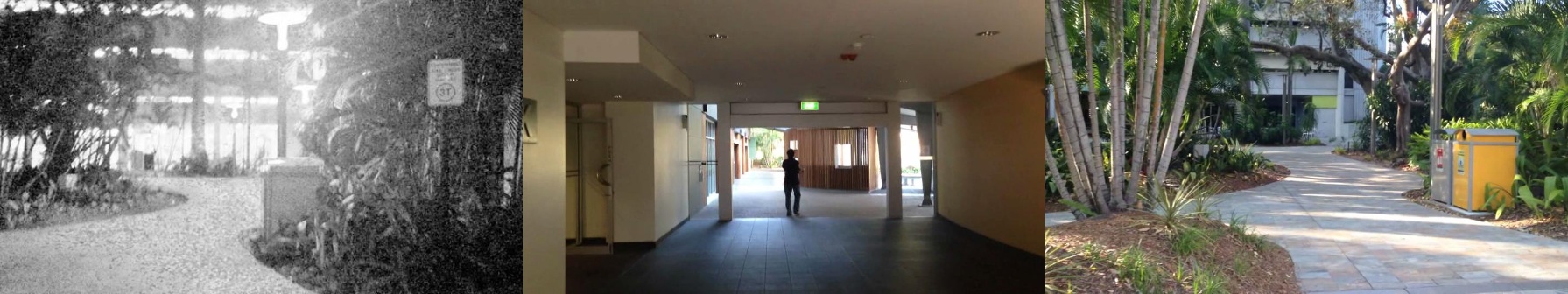}
    }

    \subfloat[Nordland]{
    \includegraphics[width=0.49\linewidth]{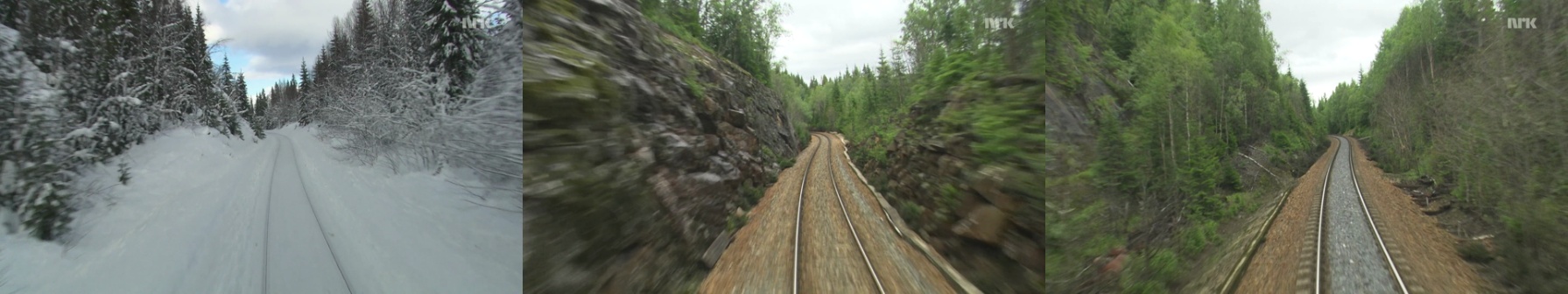}
    \hfill
    \includegraphics[width=0.49\linewidth]{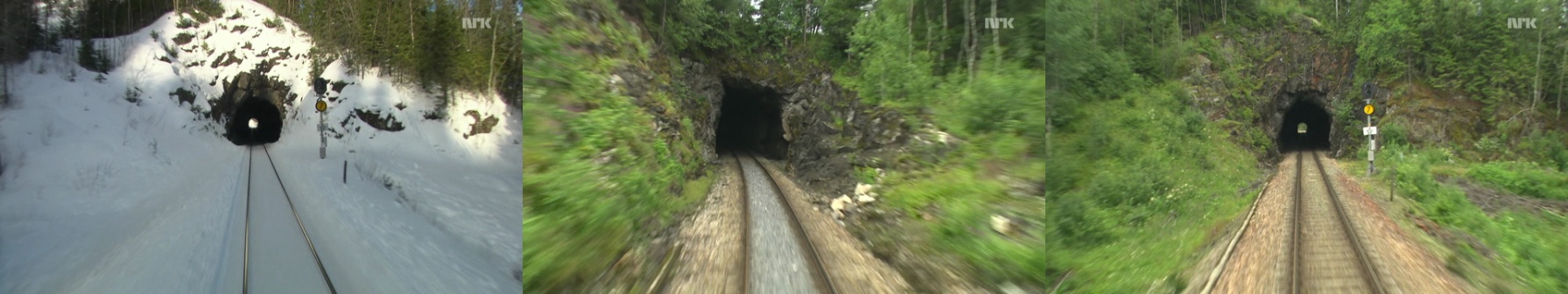}
    }

    \subfloat[Oxford RobotCar - Night]{
    \includegraphics[width=0.49\linewidth]{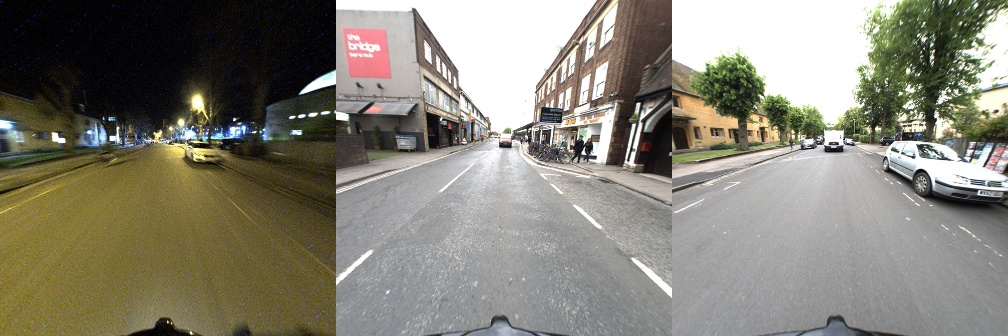}
    \hfill
    \includegraphics[width=0.49\linewidth]{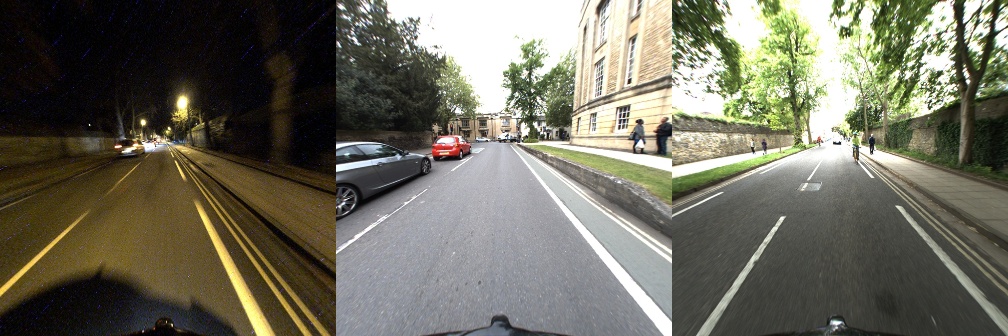}
    }

    \subfloat[Oxford RobotCar - Night-rain]{
    \includegraphics[width=0.32\linewidth]{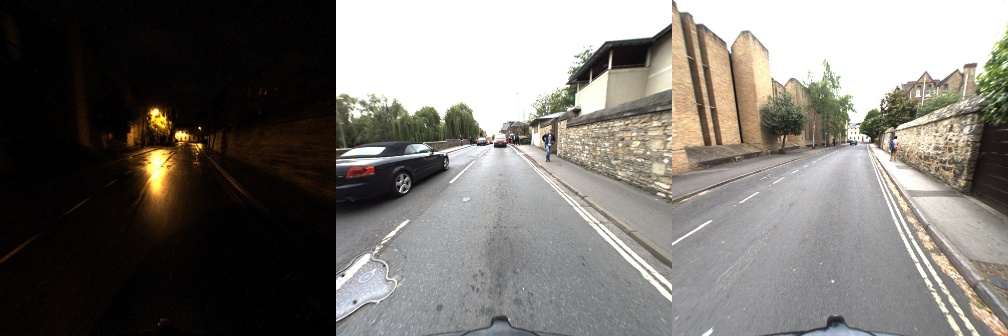}
    \hfill
    \includegraphics[width=0.32\linewidth]{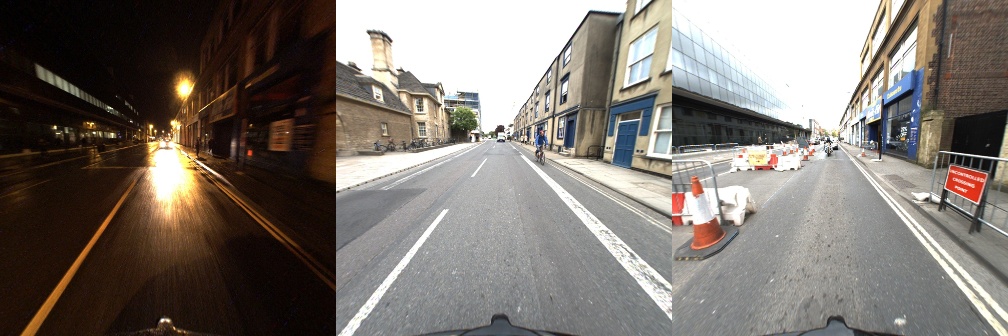}
    }
    \subfloat[Oxford RobotCar - Overcast-winter]{
    \includegraphics[width=0.32\linewidth]{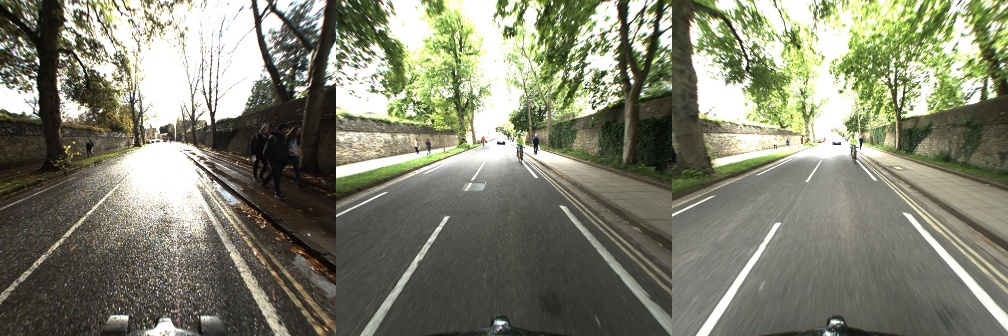}
    }

    \caption{Examples of failure cases for Mix-hist with 50 candidates and tolerance range 5. In these cases, the recognized image score is higher than the ground truth image score, so the failure is not caused by the filtering stage. Image organization: LEFT = query image, MIDDLE = recognized image, RIGHT = groud truth image\\
    The Berlin Halen. failures are caused by significant viewpoint differences. The Overcase-winter case is an example of repetitive structure. The second case from Garden Points is the only occurrence of no visible similarity between images. In all other cases, the ground truth and recognized images have similar patterns. The rest of Oxford RobotCar sequences do not contain any such failure case.}
    \label{fig:fail_cases}
\end{figure*}

Detailed results are presented in Tab.~\ref{Tab1} and for individual sequences of the Oxford RobotCar dataset in Tab. \ref{Tab2}. Average recall in all datasets for tolerance range 2 is depicted in Fig.~\ref{fig:sum}. This average includes the three Berlin datasets together (as the average of their recalls)  to balance different sizes of datasets. Finally, Fig. \ref{fig:fail_cases} on the last page shows examples of Mix-hist failures from all datasets.

Besides the D2-net feature detector, the ALIKED feature detector \cite{ALIKED} was also employed in the Mix-hist system and tested. However, these results introduced in Appendix \ref{App-aliked} were not comparable with D2-net.

All used datasets simulate scenarios with a fixed forward-looking camera mounted on a vehicle driving on flat ground, so the camera rotations are suppressed. In Appendix \ref{App-pitts}, Mix-hist and Mix-hist-comb systems were tested on a different type of dataset, Pittsburgh30k \cite{Pitts30k}. Pittsburgh30k is a standard dataset from an urban environment widely used for VPR testing. Compared with the used datasets, it contains multiple views for each location, including many images capturing only fragments of buildings or vegetation. There are also larger differences in viewpoints on the same places. On the other hand, it contains only daytime images without noticeable weather or seasonal changes.

\begin{table*}[ht]
      \caption{Recall@1 [\%] of presented systems and related methods on datasets from Sect. \ref{Datasets}. The tolerance is the allowed distance of a result from the ground truth in the reference image sequence.
      }
    \centering
    \renewcommand{\arraystretch}{1.3}
    \setlength{\tabcolsep}{5pt}
    \begin{tabular}{|l|rrr|rrr|rrr|rrr|rrr|rrr|}
    \hline 
    Dataset & \multicolumn{3}{|c|}{Berlin A100} & \multicolumn{3}{|c|}{Berlin Kudamm} & \multicolumn{3}{|c|}{Berlin Halen.} & \multicolumn{3}{|c|}{Gardens Points} & \multicolumn{3}{|c|}{Nordland} & \multicolumn{3}{|c|}{Oxford RobotCar}\\ \hline
    Tolerance & \multicolumn{1}{|c}{1} & \multicolumn{1}{c}{2} & \multicolumn{1}{c|}{5} & \multicolumn{1}{|c}{1} & \multicolumn{1}{c}{2} & \multicolumn{1}{c|}{5} & \multicolumn{1}{|c}{1} & \multicolumn{1}{c}{2} & \multicolumn{1}{c|}{5} & \multicolumn{1}{|c}{1} & \multicolumn{1}{c}{2} & \multicolumn{1}{c|}{5} & \multicolumn{1}{|c}{1} & \multicolumn{1}{c}{2} & \multicolumn{1}{c|}{5} & \multicolumn{1}{|c}{1} & \multicolumn{1}{c}{2} & \multicolumn{1}{c|}{5}\\ \hhline{|=|===|===|===|===|===|===|}
    Patch-NV perf. & 87.7 & 87.7 & 93.8 & 70.3 & 75.7 & 79.3 & 72.0 & 75.2 & 81.5 & 68.5 & 87.0 & 96.0 & 75.3 & 78.6 & 80.9 & 86.7 & 93.1 & 96.7\\ \hline
    Patch-NV speed & 90.1 & 90.1 & 95.1 & 59.9 & 67.1 & 69.8 & 66.9 & 72.6 & 78.3 & 55.0 & 66.5 & 80.5 & 60.2 & 64.1 & 67.4 & 76.4 & 86.7 & 93.1\\ \hline
    MixVPR & \textbf{97.5} & \textbf{98.8} & \textbf{98.8} & 77.0 & 85.1 & 86.9 & 85.4 & 87.3 & 89.8 & 55.0 & 69.5 & 82.0 & 89.2 & 91.7 & 92.9 & 81.6 & 90.7 & 95.9\\ \hline
    SelaVPR pits. & 96.3 & 96.3 & \textbf{98.8} & 74.8 & 81.5 & 84.2 & 82.8 & 86.0 & 89.8 & 73.5 & 88.5 & 97.5 & 92.3 & 95.4 & 96.9 & 88.9 & 95.0 & \textbf{99.4}\\ \hline
    SelaVPR msls & \textbf{97.5} & 97.5 & \textbf{98.8} & 78.8 & 86.0 & \textbf{89.2} & 86.6 & \textbf{90.4} & \textbf{94.3} & 71.0 & 88.0 & 97.5 & 91.0 & 94.5 & 95.9 & 90.4 & 95.6 & 99.1\\ \hline
    SSM-VPR & 80.2 & 84.0 & 92.6 & 75.7 & 82.9 & 84.2 & 72.0 & 75.2 & 76.4 & \textbf{83.5} & 92.0 & 92.5 & 91.7 & 92.6 & 92.9 & 89.4 & 91.8 & 93.4\\ \hline
    SSM-RAN 8 px & 75.3 & 77.8 & 81.5 & 72.1 & 80.6 & 82.0 & 73.2 & 79.0 & 80.9 & 72.0 & 90.5 & 97.0 & 94.8 & 96.2 & 96.6 & 88.1 & 92.6 & 94.8\\ \hline
    SSM-RAN 24 px & 82.7 & 86.4 & 90.1 & 74.8 & 82.0 & 84.7 & 77.7 & 82.8 & 84.1 & 72.0 & 93.0 & \textbf{98.5} & 94.0 & 95.4 & 95.9 & 86.6 & 91.8 & 94.7\\ \hline
    \hhline{|=|===|===|===|===|===|===|}
    SSM-hist & 84.0 & 87.7 & 90.1 & 73.0 & 78.8 & 80.2 & 72.0 & 76.4 & 78.3 & 79.5 & \textbf{95.0} & 97.5 & 95.1 & 96.2 & 96.6 & 89.4 & 92.5 & 94.9\\ \hline
    SSM-anchor & 76.5 & 81.5 & 84.0 & 71.6 & 77.0 & 77.9 & 68.2 & 72.0 & 73.2 & 79.0 & 91.5 & 96.5 & 93.0 & 95.1 & 95.6 & 86.4 & 90.9 & 93.5\\ \hline
    SSM-sum & 76.5 & 82.7 & 87.7 & 64.4 & 68.9 & 69.8 & 61.1 & 63.7 & 66.9 & 65.0 & 82.0 & 90.0 & 89.0 & 91.2 & 91.8 & 85.8 & 90.3 & 93.4\\ \hline
    \hhline{|=|===|===|===|===|===|===|}
    Mix-hist 10 & 90.1 & 92.6 & 97.5 & 80.6 & 86.9 & 87.8 & 80.9 & 82.8 & 84.1 & 78.0 & 90.5 & 94.5 & 96.5 & 97.9 & 98.3 & 92.1 & 96.4 & 98.4\\ \hline
    Mix-hist 50 & 84.0 & 87.7 & 91.4 & 78.4 & 85.6 & 86.9 & 76.4 & 79.0 & 80.9 & 80.0 & 92.0 & 96.0 & \textbf{97.5} & \textbf{98.5} & \textbf{98.9} & 92.8 & 96.9 & 98.5\\ \hline
    Mix-hist-comb 10 & 95.1 & 97.5 & \textbf{98.8} & \textbf{82.0} & \textbf{88.7} & \textbf{89.2} & \textbf{87.3} & 87.9 & 89.2 & 75.0 & 87.5 & 93.0 & 95.8 & 97.7 & 98.1 & 92.6 & 96.5 & 98.2\\ \hline
    Mix-hist-comb 50 & 95.1 & 97.5 & \textbf{98.8} & \textbf{82.0} & \textbf{88.7} & \textbf{89.2} & \textbf{87.3} & 87.9 & 89.2 & 78.5 & 90.0 & 95.0 & 96.3 & 98.0 & 98.4 & \textbf{93.6} & \textbf{97.4} & 98.9\\ \hline
    \end{tabular}
  \label{Tab1}
\end{table*}

\begin{table*}[ht]
    \caption{Recall@1 [\%] of presented systems and related methods (similarly to Tab. \ref{Tab1}) on various sequences of RobotCar dataset \cite{RobotCar1, RobotCar2}
    }
    \centering
    \renewcommand{\arraystretch}{1.3}
    \setlength{\tabcolsep}{2.7pt}
    \begin{tabular}{|l|rrr|rrr|rrr|rrr|rrr|rrr|rrr|}
    \hline
    Dataset & \multicolumn{3}{|c|}{Dawn} & \multicolumn{3}{|c|}{Dusk} & \multicolumn{3}{|c|}{Night} & \multicolumn{3}{|c|}{Night-rain} & \multicolumn{3}{|c|}{Overcast-winter} & \multicolumn{3}{|c|}{Rain} & \multicolumn{3}{|c|}{Snow}\\
    \hline
    Tolerance range & \multicolumn{1}{|c}{1} & \multicolumn{1}{c}{2} & \multicolumn{1}{c|}{5} & \multicolumn{1}{|c}{1} & \multicolumn{1}{c}{2} & \multicolumn{1}{c|}{5} & \multicolumn{1}{|c}{1} & \multicolumn{1}{c}{2} & \multicolumn{1}{c|}{5} & \multicolumn{1}{|c}{1} & \multicolumn{1}{c}{2} & \multicolumn{1}{c|}{5} & \multicolumn{1}{|c}{1} & \multicolumn{1}{c}{2} & \multicolumn{1}{c|}{5} & \multicolumn{1}{|c}{1} & \multicolumn{1}{c}{2} & \multicolumn{1}{c|}{5} & \multicolumn{1}{|c}{1} & \multicolumn{1}{c}{2} & \multicolumn{1}{c|}{5
}\\ \hhline{|=|===|===|===|===|===|===|===|}
    Patch-NV perf. & 89.4 & 95.9 & 98.6 & 97.8 & 99.4 & \textbf{100.0} & 65.6 & 78.7 & 87.4 & 73.9 & 84.4 & 92.7 & 93.3 & 96.9 & 97.9 & 93.8 & 98.4 & \textbf{100.0} & 93.4 & 97.8 & \textbf{100.0}\\ \hline
    Patch-NV speed & 86.7 & 94.5 & 98.2 & 90.5 & 95.5 & \textbf{100.0} & 45.4 & 60.7 & 74.3 & 48.2 & 68.8 & 82.1 & 90.8 & 96.9 & 99.0 & 88.0 & 96.4 & 98.4 & 85.0 & 94.2 & 99.6\\ \hline
    MixVPR & 85.8 & 98.2 & \textbf{100.0} & 93.9 & 99.4 & \textbf{100.0} & 53.0 & 69.4 & 83.1 & 60.1 & 75.7 & 89.4 & 92.3 & 95.9 & \textbf{100.0} & 94.8 & 98.4 & \textbf{100.0} & 91.6 & 97.8 & 98.7\\ \hline
    SelaVPR pits. & 90.8 & 95.9 & 99.5 & 95.0 & 98.3 & 99.4 & 79.2 & 86.3 & \textbf{98.9} & 81.2 & 91.7 & 98.6 & 93.3 & 97.9 & 99.5 & 90.6 & 97.9 & \textbf{100.0} & 92.5 & 96.9 & \textbf{100.0}\\ \hline
    SelaVPR msls & 91.7 & 97.7 & \textbf{100.0} & 96.1 & 98.9 & 99.4 & 83.1 & 89.1 & 98.4 & 80.7 & 87.6 & 96.3 & 93.8 & 99.0 & \textbf{100.0} & 93.2 & 99.0 & \textbf{100.0} & 94.2 & 97.8 & 99.6\\ \hline
    SSM-VPR & 93.1 & 95.0 & 97.2 & \textbf{100.0} & \textbf{100.0} & \textbf{100.0} & \textbf{85.2} & 88.0 & 90.2 & 59.6 & 62.4 & 67.4 & \textbf{96.4} & 99.0 & 99.0 & 97.4 & \textbf{100.0} & \textbf{100.0} & 94.2 & 98.2 & \textbf{100.0}\\ \hline
    SSM-RAN 8 px & 89.9 & 93.6 & 97.2 & 98.3 & 99.4 & \textbf{100.0} & 78.7 & 85.8 & 90.2 & 64.2 & 72.0 & 76.1 & 95.4 & \textbf{99.5} & \textbf{100.0} & 97.4 & 99.5 & \textbf{100.0} & 92.9 & 98.7 & \textbf{100.0}\\ \hline
    SSM-RAN 24 px & 87.6 & 91.7 & 96.8 & 98.3 & 99.4 & \textbf{100.0} & 76.0 & 85.8 & 91.8 & 61.0 & 68.8 & 74.3 & 94.4 & 99.0 & \textbf{100.0} & 95.8 & 99.0 & \textbf{100.0} & 92.9 & 99.1 & \textbf{100.0}\\ \hline
    \hhline{|=|===|===|===|===|===|===|===|}
    SSM-hist & 91.7 & 95.4 & 97.7 & 98.9 & \textbf{100.0} & \textbf{100.0} & 83.1 & 86.9 & 92.3 & 65.6 & 70.2 & 74.3 & 93.3 & 97.4 & \textbf{100.0} & 97.4 & 99.0 & \textbf{100.0} & 96.0 & 98.7 & \textbf{100.0}\\ \hline
    SSM-anchor & 87.2 & 93.1 & 96.8 & 96.1 & 99.4 & \textbf{100.0} & 74.9 & 80.3 & 86.9 & 60.6 & 67.4 & 71.1 & 93.3 & 96.9 & \textbf{100.0} & \textbf{98.4} & \textbf{100.0} & \textbf{100.0} & 94.2 & 99.1 & 99.6\\ \hline
    SSM-sum & 89.4 & 92.7 & 96.8 & 97.2 & \textbf{100.0} & \textbf{100.0} & 72.7 & 78.1 & 84.7 & 59.2 & 65.6 & 72.5 & 91.3 & 96.9 & \textbf{100.0} & 97.4 & \textbf{100.0} & \textbf{100.0} & 93.4 & 98.7 & 99.6\\ \hline
    \hhline{|=|===|===|===|===|===|===|===|}
    Mix-hist 10 & \textbf{95.0} & \textbf{98.6} & \textbf{100.0} & 97.8 & \textbf{100.0} & \textbf{100.0} & 76.5 & 84.7 & 90.2 & \textbf{91.7} & \textbf{97.2} & \textbf{99.1} & 90.8 & 95.9 & 99.5 & 97.4 & 99.0 & \textbf{100.0} & 95.6 & 99.1 & \textbf{100.0}\\ \hline
    Mix-hist 50 & 94.5 & 98.2 & 99.5 & 97.8 & \textbf{100.0} & \textbf{100.0} & 82.0 & 89.6 & 93.4 & \textbf{91.7} & 96.3 & 96.8 & 90.8 & 95.9 & 99.5 & 97.4 & 99.0 & \textbf{100.0} & 95.6 & 99.1 & \textbf{100.0}\\ \hline
    Mix-hist-comb 10 & 93.6 & \textbf{98.6} & \textbf{100.0} & 98.3 & \textbf{100.0} & \textbf{100.0} & 78.1 & 84.7 & 89.1 & 89.4 & 95.9 & 98.6 & 94.4 & 97.9 & 99.5 & 97.9 & 99.0 & \textbf{100.0} & \textbf{96.5} & \textbf{99.6} & \textbf{100.0}\\ \hline
    Mix-hist-comb 50 & 93.6 & \textbf{98.6} & \textbf{100.0} & 98.3 & \textbf{100.0} & \textbf{100.0} & 83.6 & \textbf{90.2} & 94.0 & 90.8 & 96.3 & 98.6 & 94.4 & 97.9 & 99.5 & 97.9 & 99.0 & \textbf{100.0} & \textbf{96.5} & \textbf{99.6} & \textbf{100.0}\\ \hline
    \end{tabular}
  \label{Tab2}
\end{table*}

\subsection{Computational Complexity}
\label{Complexity}
Besides the VPR precision, the computational complexity for all the investigated methods was compared. The experiments were performed on a computer equipped with Intel Core i7-7700 processor, Nvidia GeForce RTX 2080 Ti graphic card, and 64 GB of RAM.

The first measured characteristic is the size of the database storing visual features of the reference sequence. The values for Oxford RobotCar and Nordland datasets are summarized in Tab.~\ref{Tab_memory}.

\begin{figure}[t]
    \centering
    \includegraphics[width=\columnwidth]{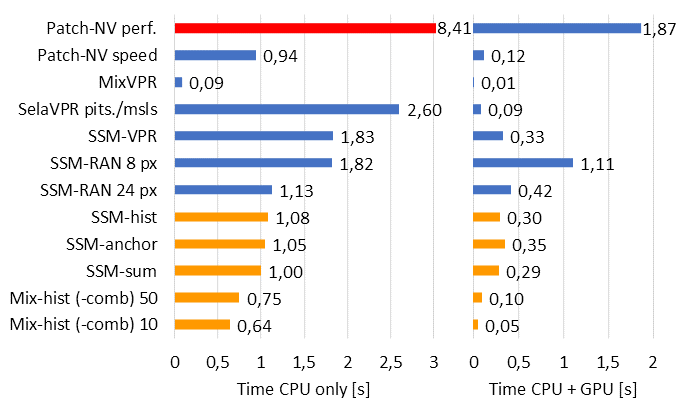}
    \caption{Average VPR computational time per one query frame for the Oxford RobotCar Snow sequence.}    
    \label{fig:speed}
\end{figure}

The second measured property is the average computation time required for single query image recognition. The values measured on the Oxford RobotCar Snow sequence are depicted in Fig. \ref{fig:speed}. As the presented methods are primarily designed for robotic applications, a GPU unit must not always be available. Therefore, the time was measured separately for CPU only and CPU with GPU support. Additionally, the computation times for individual steps of all introduced systems, SSM-VPR and SSM-RAN, using GPU, are presented in Tab. \ref{Tab_stepstime}. Since the number of input candidates for the re-ranking stage is fixed here, the computational time does not depend on the number of reference images.

\begin{table}[t]
    \centering
    \caption{Sizes of reference databases for selected datasets}
    \renewcommand{\arraystretch}{1.3}
    \begin{tabular}{|l|r|r|}
    \hline
     & RobotCar & Nordland\\ \hline
    Patch-NV perf. & 7 040 MB & 62 520 MB \\ \hline
    Patch-NV speed & 292 MB & 2 596 MB \\ \hline
    MixVPR & 7 MB & 45 MB \\ \hline
    SelaVPR & 401 MB & 2 577 MB \\ \hline
    SSM-VPR & 163 MB & 913 MB \\ \hline
    SSM-hist, anchor, sum, RANSAC & 344 MB & 2 120 MB \\ \hline
    Mix-hist & 304 MB & 1 942 MB \\ \hline
    \end{tabular}
  \label{Tab_memory}
\end{table}

\begin{table}[t]
    \centering
    \caption{Computation time [s] of individual steps of VPR using GPU}
    \renewcommand{\arraystretch}{1.3}
    \begin{tabular}{|l|c|c|c|c|c|c|}
    \hline
     & \makecell[c]{ Mix \\ hist \\ 10 / 50 } &  \makecell[c]{ SSM \\ hist } & \makecell[c]{ SSM \\ anchor } & \makecell[c]{ SSM \\ sum } & \makecell[c]{ SSM \\ RAN \\ 8 / 24 px } & \makecell[c]{ SSM \\ VPR } \\ \hline
    Stage I & 0.01 & \multicolumn{5}{c|}{0.18}\\ \hline
    \makecell[l]{ Feature \\ Detection } & 0.03 & \multicolumn{4}{c|}{0.05} & 0.01\\ \hline
    \makecell[l]{ Matching + \\ Score } & \makecell[c]{ 0.01 \\ 0.07 } &0.07 & 0.11 & 0.06 & \makecell[c]{ 0.88 \\ 0.19 } & 0.14\\
    \hhline{|=|=|=|=|=|=|=|}
    Total & \makecell[c]{ 0.05 \\ 0.10 } & 0.30 & 0.35 & 0.29 & \makecell[c]{ 1.11 \\ 0.42 } & 0.33\\ \hline
    \end{tabular}

  \label{Tab_stepstime}
\end{table}

\subsection{Discussion of Results}
\label{Discussion}
\subsubsection{Mutual comparison of the introduced methods}
All presented methods with the D2-net detector work well on both urban and non-urban datasets and evince high robustness to seasonal, weather, and daytime changes. The SSM-hist achieved the best results from these methods. In addition, its speed is comparable to the simplest and fastest method, SSM-sum, which exhibits significantly lower precision. The SSM-anchor method achieves precision similar to SSM-hist but at a lower speed using GPU, so in most applications, SSM-hist appears more convenient. 
Therefore, the histogram of shifts method was selected for further testing of the system with the MixVPR filtering stage.

\subsubsection{Comparison with SSM-VPR and SSM-RAN}
Compared with the original SSM-VPR, SSM-hist achieves slightly higher average precision and especially excels on the Nordland dataset. For the Gardens Points dataset, SSM-hist is less precise, so it provides lower recall for the lowest threshold. By contrast, it outperforms the SSM-VPR for higher thresholds. The experiments confirmed that the presented methods combined with the D2-net detector achieve results and computation time comparable to the original SSM-VPR approach. Although the D2-net feature database has approximately doubled memory requirements, it still remains manageable even for larger datasets, so the applicability of the presented system is not significantly limited.

SSM-RAN performs slightly better on Berlin datasets containing larger rotations and lateral shifts, on which it benefits from a more general model. On the other hand, SSM-hist achieves higher overall precision with lower computational cost and confirms the benefits of model-free methods. The SSM-RAN version with a larger re-projection threshold 24 px, which mitigates the conditions for model inlier acceptance, achieves better results than SSM-RAN with the 8px threshold. In addition, its RANSAC confidence increases more quickly, which significantly reduces the number of iterations and computational time.

\subsubsection{Comparison with Patch-NetVLAD}

The Patch-NetVLAD perf. version reaches comparable results as the SSM-hist system on Berlin datasets and Oxford RobotCar, but it has been significantly overcome on Garden Points and Nordland datasets. In addition, the computation time and database size for the Patch-NV perf. version are multiple times larger than for the other systems. The Patch-NV speed version is significantly faster, especially with the use of GPU, and demands less memory. Nevertheless, its performance is the worst of all tested methods. In comparison with solutions based on SSM-VPR, both Patch-NetVLAD systems excel on the Night-rain sequence from the Oxford RobotCar dataset. However, this is primarily caused by the exceptional failure of the SSM-VPR filtering stage, which does not return any correct candidate in 23 \% of cases.

\subsubsection{Comparison with MixVPR and SelaVPR}
MixVPR excels especially on all Berlin datasets, but the SSM-hist outperforms it on the remaining datasets. On the other hand, the substantial advantage of MixVPR is its computational efficiency, which can be essential for real-time applications.

SelaVPR also performs well on Berlin datasets. Moreover, it reaches comparable results as SSM-hist on other datasets and has better overall average precision.

\subsubsection{Mix-hist and Mix-hist-comb}
Since MixVPR and SelaVPR significantly outperformed the SSM-hist approach on some datasets, the histogram of shifts method with the MixVPR filtering stage was further tested. This system achieved even better overall results than SelaVPR, and the re-ranking significantly enhances the MixVPR precision on most datasets. The memory requirements and computational time with the use of GPU are comparable to those of SelaVPR, and in addition, the Mix-hist system is more efficient with the use of CPU only. From the two tested versions differing by the used number of candidates, Mix-hist 50 performs better on Gardens Points, Nordland, and Oxford RobotCar datasets, but Mix-hist 10 exhibits slightly better overall precision.

The combination of filtering and re-ranking scores introduced in Sect. \ref{MixVPR-version} improves the overall performance, and Mix-hist-comb-50 achieves the best results from all systems. On the other hand, standalone re-ranking still performs better on Garden Points and Nordland datasets.

\section{Conclusions}
\label{Conclusions}
The article introduces three new model-free re-ranking approaches generally applicable to all standard local visual features. These methods were primarily designed for deep-learned local visual features since they are intended especially for long-term autonomy applications. The crucial advantage of this type of features is that they generally maintain high robustness to various appearance changes in the environment.

For evaluation and testing of the introduced approaches, they were implemented with a D2-net feature detector \cite{D2-Net} into a new VPR system. It combines the proposed re-ranking methods with SSM-VPR filtering stage \cite{SSM-VPR} or MixVPR \cite{MixVPR}. The system was experimentally evaluated on six public datasets and directly compared with several state-of-the-art solutions. 
The experiments confirmed that the introduced model-free methods are well-suited for long-term VPR. The histogram of shifts re-ranking method with newer filtering achieves the best results on the Nordland dataset and the highest average recall from all tested methods. In addition, the Mix-hist-comb version demonstrated that the re-ranking performance can benefit from filtering stage scores.

The general applicability of the new methods to all standard local visual features opens the possibility of using them with other robust deep-learned detectors and reaching new state-of-the-art results for long-term VPR. In addition, the new VPR system and the introduced methods can be directly employed in teach-and-repeat navigation similarly to \cite{SSM-TaR} or other localization systems (e.g., SLAM).

\appendices
\section{ALIKED Feature Detector}
\label{App-aliked}
The histogram of shifts method with the ALIKED \cite{ALIKED} detector and MixVPR filtering stage was also tested. ALIKED detector belongs among current state-of-the-art methods for various computer vision tasks such as 3D reconstruction or image matching. For testing, the original aliked-n32 pre-trained version was used. The image size was 320 x 320 px. In comparison with the original implementation, only the score threshold for feature detection was lowered to 0.1 to obtain more keypoints. Nevertheless, the achieved preliminary results presented in Tab. \ref{Tab_aliked} were significantly worse than those for D2-net, so the ALIKED detector in the used configuration is not competitive.

\begin{table}[t]
    \centering
    \caption{Recall@1 [\%] of Mix-hist method with ALIKED detector \cite{ALIKED} on datasets from Sect. \ref{Datasets}.}
    \renewcommand{\arraystretch}{1.3}
    \begin{tabular}{|l|r|r|r|}
    \hline
    Tolerance range & \multicolumn{1}{c}{1} & \multicolumn{1}{|c}{2} & \multicolumn{1}{|c|}{5} \\ \hline
    Berlin A100 & 71.6 & 74.1 & 79.0\\ \hline
    Berlin Kudamm & 60.8 & 68.5 & 69.8\\ \hline
    Berlin Halen. & 63.1 & 67.5 & 72.0\\ \hline
    Gardens Points & 67.5 & 81.5 & 88.5\\ \hline
    Nordland & 83.0 & 85.4 & 87.4\\ \hline
    Oxford RobotCar & 85.2 & 90.3 & 92.5\\ \hline
    \end{tabular} 
  \label{Tab_aliked}
\end{table}

\section{Pittsburgh30k Dataset}
\label{App-pitts}
Tab. \ref{Tab_pitts30k} shows the comparison of Mix-hist and Mix-hist-comb with state-of-the-art methods on standard Pittsburgh30k dataset widely used for VPR benchmarking. The experiments were performed on the testing subset of the Pittsburgh30k dataset with 10,000 database images. The used metrics are various recalls measuring the percentage of cases with at least one correct match between the selected number of the best candidates. The match is considered correct if the real distance of the found location from the ground truth is beyond 25 m.

The achieved results are comparable with the current state-of-the-art methods, and the re-ranking increases the performance of MixVPR in most cases. The only exception is Mix-hist with 50 candidates, slightly decreasing the original recall@1. SelaVPR outperforms the proposed methods; however, these methods were not designed or trained for this dataset type.

\begin{table}[b]
    \centering
    \caption{Recalls of Mix-hist and Mix-his-comb systems on Pittsburgh30k dataset\cite{Pitts30k}}
    \renewcommand{\arraystretch}{1.3}
    \begin{tabular}{|l|c|c|c|}
    \hline
     & Recall@1 & Recall@5 & Recall@10 \\ \hline
    MixVPR & 90.8 & 97.1 & 98.5\\ \hline
    SelaVPR pits. stage 1 & 94.2 & \textbf{99.4} & 99.7\\ \hline
    SelaVPR pits. & \textbf{94.7} & 99.3 & \textbf{99.8}\\ \hline
    SelaVPR msls stage 1 & 86.9 & 96.7 & 98.3\\ \hline
    SelaVPR msls & 91.5 & 98.3 & 99.0\\ \hline \hline
    Mix-hist 10 & 91.5 & 97.5 & 98.5\\ \hline
    Mix-hist 50 & 90.4 & 97.4 & 98.7\\ \hline
    Mix-hist-comb 10 & 92.1 & 97.5 & 98.5\\ \hline
    Mix-hist-comb 50 & 92.1 & 97.6 & 98.6\\ \hline
    \end{tabular} 
  \label{Tab_pitts30k}
\end{table}

\pagebreak

\bibliographystyle{IEEEtran}
\bibliography{bibliography}

\begin{thebibliography}{10}
\providecommand{\url}[1]{#1}
\csname url@samestyle\endcsname
\providecommand{\newblock}{\relax}
\providecommand{\bibinfo}[2]{#2}
\providecommand{\BIBentrySTDinterwordspacing}{\spaceskip=0pt\relax}
\providecommand{\BIBentryALTinterwordstretchfactor}{4}
\providecommand{\BIBentryALTinterwordspacing}{\spaceskip=\fontdimen2\font plus
\BIBentryALTinterwordstretchfactor\fontdimen3\font minus \fontdimen4\font\relax}
\providecommand{\BIBforeignlanguage}[2]{{%
\expandafter\ifx\csname l@#1\endcsname\relax
\typeout{** WARNING: IEEEtran.bst: No hyphenation pattern has been}%
\typeout{** loaded for the language `#1'. Using the pattern for}%
\typeout{** the default language instead.}%
\else
\language=\csname l@#1\endcsname
\fi
#2}}
\providecommand{\BIBdecl}{\relax}
\BIBdecl

\bibitem{VPR_survey}
C.~Masone and B.~Caputo, ``A survey on deep visual place recognition,'' \emph{IEEE Access}, vol.~9, pp. 19\,516--19\,547, 2021.

\bibitem{SSM-TaR}
L.~G. Camara, T.~Pivo\v{n}ka, M.~J\'ilek, C.~G\"abert, K.~Ko\v{s}nar, and L.~P\v{r}eu\v{c}il, ``Accurate and robust teach and repeat navigation by visual place recognition: A cnn approach,'' in \emph{2020 IEEE/RSJ International Conference on Intelligent Robots and Systems (IROS)}, 2020, pp. 6018--6024.

\bibitem{Pairwise_match}
X.~Li, M.~Larson, and A.~Hanjalic, ``Pairwise geometric matching for large-scale object retrieval,'' in \emph{2015 IEEE Conference on Computer Vision and Pattern Recognition (CVPR)}, 2015, pp. 5153--5161.

\bibitem{Patch-NetVLAD}
S.~Hausler, S.~Garg, M.~Xu, M.~Milford, and T.~Fischer, ``Patch-netvlad: Multi-scale fusion of locally-global descriptors for place recognition,'' in \emph{2021 IEEE/CVF Conference on Computer Vision and Pattern Recognition (CVPR)}, 2021, pp. 14\,136--14\,147.

\bibitem{Zhang}
Y.~Zhang, Z.~Jia, and T.~Chen, ``Image retrieval with geometry-preserving visual phrases,'' in \emph{2011 IEEE Conference on Computer Vision and Pattern Recognition (CVPR)}, 2011, pp. 809--816.

\bibitem{SSM-VPR}
L.~G. Camara and L.~P\v{r}eu\v{c}il, ``Visual place recognition by spatial matching of high-level cnn features,'' \emph{Robotics and Autonomous Systems}, vol. 133, p. 103625, 2020.

\bibitem{D2-Net}
M.~Dusmanu, I.~Rocco, T.~Pajdla, M.~Pollefeys, J.~Sivic, A.~Torii, and T.~Sattler, ``D2-net: A trainable cnn for joint description and detection of local features,'' in \emph{2019 IEEE/CVF Conference on Computer Vision and Pattern Recognition (CVPR)}, 2019, pp. 8084--8093.

\bibitem{local-features-for-VPR}
G.~Barbarani, M.~Mostafa, H.~Bayramov, G.~Trivigno, G.~Berton, C.~Masone, and B.~Caputo, ``Are local features all you need for cross-domain visual place recognition?'' in \emph{2023 IEEE/CVF Conference on Computer Vision and Pattern Recognition Workshops (CVPRW)}, 2023, pp. 6155--6165.

\bibitem{MixVPR}
A.~Ali-Bey, B.~Chaib-Draa, and P.~Giguére, ``Mixvpr: Feature mixing for visual place recognition,'' in \emph{2023 IEEE/CVF Winter Conference on Applications of Computer Vision (WACV)}, 2023, pp. 2997--3006.

\bibitem{VLAD}
H.~Jégou, M.~Douze, C.~Schmid, and P.~Pérez, ``Aggregating local descriptors into a compact image representation,'' in \emph{2010 IEEE Computer Society Conference on Computer Vision and Pattern Recognition}, 2010, pp. 3304--3311.

\bibitem{NetVLAD}
R.~Arandjelovic, P.~Gronat, A.~Torii, T.~Pajdla, and J.~Sivic, ``Netvlad: Cnn architecture for weakly supervised place recognition,'' in \emph{2016 IEEE Conference on Computer Vision and Pattern Recognition (CVPR)}, 2016, pp. 5297--5307.

\bibitem{HDME}
N.~V. Keetha, M.~Milford, and S.~Garg, ``A hierarchical dual model of environment- and place-specific utility for visual place recognition,'' \emph{IEEE Robotics and Automation Letters}, vol.~6, no.~4, pp. 6969--6976, 2021.

\bibitem{SuperPoint}
D.~DeTone, T.~Malisiewicz, and A.~Rabinovich, ``Superpoint: Self-supervised interest point detection and description,'' in \emph{2018 IEEE/CVF Conference on Computer Vision and Pattern Recognition Workshops (CVPRW)}, 2018, pp. 337--33\,712.

\bibitem{SuperGlue}
P.-E. Sarlin, D.~DeTone, T.~Malisiewicz, and A.~Rabinovich, ``Superglue: Learning feature matching with graph neural networks,'' in \emph{2020 IEEE/CVF Conference on Computer Vision and Pattern Recognition (CVPR)}, 2020, pp. 4937--4946.

\bibitem{GeM}
F.~Radenović, G.~Tolias, and O.~Chum, ``Fine-tuning cnn image retrieval with no human annotation,'' \emph{IEEE Transactions on Pattern Analysis and Machine Intelligence}, vol.~41, no.~7, pp. 1655--1668, 2019.

\bibitem{CosPlace}
G.~Berton, C.~Masone, and B.~Caputo, ``Rethinking visual geo-localization for large-scale applications,'' in \emph{2022 IEEE/CVF Conference on Computer Vision and Pattern Recognition (CVPR)}, 2022, pp. 4868--4878.

\bibitem{SelaVPR}
F.~Lu, L.~Zhang, X.~Lan, S.~Dong, Y.~Wang, and C.~Yuan, ``Towards seamless adaptation of pre-trained models for visual place recognition,'' in \emph{The Twelfth International Conference on Learning Representations}, 2024.

\bibitem{Lowe2004}
D.~G. Lowe, ``Distinctive image features from scale-invariant keypoints,'' \emph{International Journal of Computer Vision}, vol.~60, no.~2, pp. 91--110, 2004.

\bibitem{Avrithis}
\BIBentryALTinterwordspacing
Y.~Avrithis and G.~Tolias, ``Hough pyramid matching,'' \emph{International Journal of Computer Vision}, vol. 107, no.~1, pp. 1--19, 2014. [Online]. Available: \url{http://link.springer.com/10.1007/s11263-013-0659-3}
\BIBentrySTDinterwordspacing

\bibitem{Shen}
X.~Shen, Z.~Lin, J.~Brandt, S.~Avidan, and Y.~Wu, ``Object retrieval and localization with spatially-constrained similarity measure and k-nn re-ranking,'' in \emph{2012 IEEE Conference on Computer Vision and Pattern Recognition (CVPR)}, 2012, pp. 3013--3020.

\bibitem{LoST}
S.~Garg, N.~Suenderhauf, and M.~Milford, ``Lost? appearance-invariant place recognition for opposite viewpoints using visual semantics,'' \emph{Proceedings of Robotics: Science and Systems XIV}, 2018.

\bibitem{MegaDepth}
Z.~Li and N.~Snavely, ``Megadepth: Learning single-view depth prediction from internet photos,'' in \emph{2018 IEEE/CVF Conference on Computer Vision and Pattern Recognition}, 2018, pp. 2041--2050.

\bibitem{RobotCar1}
T.~Sattler, W.~Maddern, C.~Toft, A.~Torii, L.~Hammarstrand, E.~Stenborg, D.~Safari, M.~Okutomi, M.~Pollefeys, J.~Sivic, F.~Kahl, and T.~Pajdla, ``Benchmarking 6dof outdoor visual localization in changing conditions,'' in \emph{2018 IEEE/CVF Conference on Computer Vision and Pattern Recognition (CVPR)}, 2018, pp. 8601--8610.

\bibitem{RobotCar2}
W.~Maddern, G.~Pascoe, C.~Linegar, and P.~Newman, ``{1 Year, 1000km: The Oxford RobotCar Dataset},'' \emph{The International Journal of Robotics Research (IJRR)}, vol.~36, no.~1, pp. 3--15, 2017.

\bibitem{Mapillary}
\BIBentryALTinterwordspacing
(2024) Mapillary. [Online]. Available: \url{https://www.mapillary.com}
\BIBentrySTDinterwordspacing

\bibitem{Berlin-datasets}
Z.~Chen, F.~Maffra, I.~Sa, and M.~Chli, ``Only look once, mining distinctive landmarks from convnet for visual place recognition,'' in \emph{2017 IEEE/RSJ International Conference on Intelligent Robots and Systems (IROS)}, 2017, pp. 9--16.

\bibitem{GardensPoint}
\BIBentryALTinterwordspacing
A.~Glover, ``Day and night, left and right,'' Mar. 2014. [Online]. Available: \url{https://doi.org/10.5281/zenodo.4590133}
\BIBentrySTDinterwordspacing

\bibitem{NordLand}
\BIBentryALTinterwordspacing
S.~Skrede. (2013) Nordlandsbanen: minute by minute, season by season. [Online]. Available: \url{https://nrkbeta.no/2013/01/15/nordlandsbanen-minute-by-minute-season-by-season}
\BIBentrySTDinterwordspacing

\bibitem{NordLandPaper}
P.~Neubert, N.~Sünderhauf, and P.~Protzel, ``Superpixel-based appearance change prediction for long-term navigation across seasons,'' \emph{Robotics and Autonomous Systems}, vol.~69, pp. 15--27, 2015, selected papers from 6th European Conference on Mobile Robots.

\bibitem{ALIKED}
\BIBentryALTinterwordspacing
X.~Zhao, X.~Wu, W.~Chen, P.~C.~Y. Chen, Q.~Xu, and Z.~Li, ``Aliked: A lighter keypoint and descriptor extraction network via deformable transformation,'' \emph{IEEE Transactions on Instrumentation \& Measurement}, vol.~72, pp. 1--16, 2023. [Online]. Available: \url{https://arxiv.org/pdf/2304.03608.pdf}
\BIBentrySTDinterwordspacing

\bibitem{Pitts30k}
A.~Torii, J.~Sivic, T.~Pajdla, and M.~Okutomi, ``Visual place recognition with repetitive structures,'' in \emph{2013 IEEE Conference on Computer Vision and Pattern Recognition}, 2013, pp. 883--890.

\end{thebibliography}


\section*{Biography Section}

\begin{IEEEbiography}[{\includegraphics[width=1in,height=1.25in,clip,keepaspectratio]{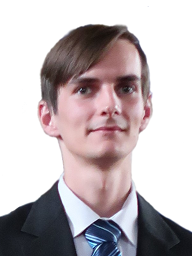}}]{Tom\'a\v{s} Pivo\v{n}ka} has received his master's degree in robotics at Faculty of Electrical Engineering of Czech Technical University in
Prague (CTU) in 2018, where he continues in Ph.D. study program Artificial Intelligence and Biocybernetics. He works at the Intelligent and Mobile Robotics Group of Czech Institute of Informatics, Robotics and Cybernetics, CTU.
His main research interests are visual localization, navigation, and computer vision.
\end{IEEEbiography}


\begin{IEEEbiography}
[{\includegraphics[width=1in,height=1.25in,clip,keepaspectratio]{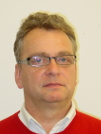}}]{Libor P\v{r}eu\v{c}il} 
Libor Přeučil received his M.Sc. in Technical Cybernetics and Robotics (1985) and Ph.D. in Computer Vision (1993), both from the Czech Technical University in Prague (CTU). He is founder and head of the Intelligent and Mobile Robotics laboratory (IMR) at the Czech Institute of Informatics, Robotics and Cybernetics (CIIRC) within the Czech Technical University in Prague. Libor Přeučil has over 20 year experience in teaching and leading R\&D on national, European and world-wide level with theoretical results and practical outcomes into industrial applications.  
His major research interests focus robot cognition and autonomy for UGV and UAV robot systems, smart manipulation, etc. This comprises sensing, localization, planning and scheduling in advanced navigation for life-long autonomy of robots in infrastructure-free, daily, natural and/or urban and human-oriented spaces. Libor Přeučil has also co-founded Center for Advanced Field Robotics re-boiling outdoor autonomy know-how into industrial-grade solutions.  
He is also internationally recognized as organizer of large scientific conferences as ECMR2019, general chair of 2021 IEEE/RSJ IROS, 2022 IFAC IAV, etc. and  (co)author of over 180 research papers, several edited books attaining H-index 19 (WoS) and 31 (Google Scholar).
\end{IEEEbiography}

\vfill

\end{document}